\documentclass[12pt]{article}
\usepackage{amsmath}
\usepackage{graphicx}
\usepackage{enumerate}
\usepackage{natbib}
\usepackage{url} 
\usepackage{amssymb}
\usepackage{amsthm}
\usepackage{float}
\usepackage{enumitem}
\usepackage{bbm,bm}
\usepackage{color}
\usepackage[dvipsnames]{xcolor}
\usepackage{caption}
\usepackage{subcaption}
\usepackage{graphicx}
\usepackage{graphics}
\usepackage[letterpaper=true,colorlinks=true,pdfpagemode=none,urlcolor=blue,linkcolor=blue,
citecolor=blue,pdfstartview=FitH,plainpages=true]{hyperref} 
\usepackage{algorithm}
\usepackage{algpseudocode}
\newcommand{\blind}{0}
\everymath{\displaystyle}
\newcommand{\bfm}[1]{\ensuremath{\mathbf{#1}}}
\def\mA{\mathcal{A}}
\def\mI{\mathcal{I}}
\def\mS{\mathcal{S}}
     
\def\bb{\bfm b}     
     
\def\bd{\bfm d}

     \def\bG{\bfm G}

\def\bv{\bfm v}     
\def\bw{\bfm w}     
\def\bx{\bfm x}     \def\bX{\bfm X}
     \def\bY{\bfm Y}
     
\def\bSigma{ \bfm \Sigma}  
\def\bepsilon{ \bfsym \epsilon}  
\def\*{\times}
\newcommand{\bfsym}[1]{\ensuremath{\boldsymbol{#1}}}

\def\btheta{\bfsym \theta}

\def\bdelta{\bfsym \delta}

\def\bSigma{\bfsym \Sigma}

\def\cA{{\cal  A}}

\def\cI{{\cal  I}}

\renewcommand{\hat}{\widehat}
\renewcommand{\tilde}{\widetilde}

\def\bSigma{\bfsym \Sigma}

\DeclareMathOperator*{\argmin}{arg\,min}  

\theoremstyle{plain}
\newtheorem{theorem}{Theorem}[section]
\newtheorem{lemma}[theorem]{Lemma}

\newtheorem{corollary}[theorem]{Corollary}

\theoremstyle{definition}

\theoremstyle{remark}

\numberwithin{equation}{section}
\begin{document}

\def\spacingset#1{\renewcommand{\baselinestretch}%
{#1}\small\normalsize} \spacingset{1}


\if0\blind
{
  \title{\bf Minimax and Communication-Efficient Distributed Best Subset Selection with Oracle Property}
  \author{Jingguo Lan\\
    University of Science and Technology of China
    \\
    Hongmei Lin \thanks{Lin’s research was partially supported by the National Natural Science Foundation of China (12171310, 12371272), the Shanghai ``Project Dawn 2022''(22SG52), and the Basic Research Project of Shanghai Science and Technology Commission (22JC1400800);}\hspace{.2cm}\\
    Shanghai University of International Business and Economics
    and \\
    Xueqin Wang \thanks{ Wang's research is partially supported by the National Natural Science Foundation of China ( 72171216, 12231017, 71921001, and 71991474). Hongmei Lin and Xueqin Wang are co-corresponding authors.}\hspace{.2cm}\\
    University of Science and Technology of China}

  \maketitle
 \fi

\if1\blind
{
  \bigskip
  \bigskip
  \bigskip
  \begin{center}
    {\Large \bf Minimax and Communication-Efficient Distributed Best Subset Selection with Oracle Property}
\end{center}
  \medskip
} \fi

\bigskip
\begin{abstract}

The explosion of large-scale data in fields such as finance, e-commerce, and social media has outstripped the processing capabilities of single-machine systems, driving the need for distributed statistical inference methods. Traditional approaches to distributed inference often struggle with achieving true sparsity in high-dimensional datasets and involve high computational costs. We propose a novel, two-stage, distributed best subset selection algorithm to address these issues. Our approach starts by efficiently estimating the active set while adhering to the $\ell_0$ norm-constrained surrogate likelihood function, effectively reducing dimensionality and isolating key variables. A refined estimation within the active set follows, ensuring sparse estimates and matching the minimax $\ell_2$ error bound. We introduce a new splicing technique for adaptive parameter selection to tackle subproblems under $\ell_0$ constraints and a Generalized Information Criterion (GIC). Our theoretical and numerical studies show that the proposed algorithm correctly finds the true sparsity pattern, has the oracle property, and greatly lowers communication costs. This is a big step forward in distributed sparse estimation.

\end{abstract}

\noindent%
\vfill

\newpage
\spacingset{1.9} 
\section{Introduction}
\label{sec:intro}

\subsection{Literature review}

Rapid advances in information technology have generated vast datasets across domains like finance, e-commerce, and social media in the digital age. Processing such large-scale data on a single machine is often impractical due to constraints in CPU capacity, memory, and the need for privacy protection. These challenges have driven the development of new statistical inference methods tailored for distributed environments, ensuring efficient estimation while minimizing communication costs.

Recent progress in distributed statistical inference can be categorized into two main approaches. The first is the ``divide and conquer" framework, where each machine computes an estimate using its local data, and a global estimate is then obtained by averaging the estimates from different machines. This entire process requires only a single round of communication \citep{Zhang2013ComunicationEfficient, Liu2014Distributed, Gu2022Weighted, Lee2017Communicationefficient, huang2019distributed}.
The second approach involves approximating the global loss function by modifying the local loss functions. A notable example is the Distributed Approximate Newton (DANE) method proposed by \cite{Shamir2014CommunicationEfficient}. Each machine minimizes a modified loss function in this method based on its data and the gradient received from other machines during each iteration. The core idea is to reduce the discrepancy between local and global loss functions iteratively. Following this line of thinking, \citet{Chen2022FirstOrder} created a quasi-Newton algorithm that uses first-order data, and \citet{Wang2017Efficient} and \citet{Jordan2019CommunicationEfficient} created the communication-efficient surrogate likelihood (CSL) framework. This framework uses a gradient-enhanced function to get close to the global loss. The Communication-Efficient Accurate Statistical Estimation (CEASE) algorithm was created by \cite{Fan2021CommunicationEfficient} based on these methods and was inspired by the Proximal Point Algorithm (PPA). It makes the algorithm much less dependent on initial values.

Developing sparse models through distributed computing has become a major research focus in scenarios with large samples and high-dimensional features. Sparse models, which play a crucial role in traditional statistical learning, are known for efficiently handling high-dimensional datasets by selecting a sparse set of predictor variables. This property enhances the identification of key predictors, leading to more meaningful and interpretable results \citep{Tibshirani1996, Fan2001, wang2007tuning}. Recent advancements have extended these sparse models to distributed environments. For instance, \citet{Lee2017Communicationefficient} proposed the debiased Lasso technique, which aggregates debiased Lasso estimates from local machines to produce a global estimate. However, the debiased Lasso does not yield truly sparse estimates, and its error bound does not achieve the minimax $\ell_2$ rate. A truncation method is introduced to ensure sparsity, though selecting the appropriate threshold in practice is challenging \citep{Battey2018}. The distributed best subset selection method proposed by \cite{chen2023communication} also encounters similar issues. Additionally, \citet{Wang2017Efficient} and \citet{Jordan2019CommunicationEfficient} combined the CSL framework with an $\ell_1$ penalty to focus on $\ell_1$ shrinkage estimation. Despite this, the CSL method does not produce sparse estimates, and the cost of tuning the $\ell_1$ penalty parameter is prohibitively high.
Furthermore, \citet{Zhu2021LeastSquare} introduced the Distributed Least Squares (DLSA) method, which calculates least squares estimates locally and integrates them with adaptive Lasso and least angle regression techniques to achieve sparse estimations. However, the DLSA method requires the transmission of covariance matrices between machines, resulting in substantial communication costs. 

In summary, research into sparse learning or variable selection in distributed systems is still in its early stages. It faces two significant challenges \citep{Zhu2021LeastSquare}: (i) Many studies have not established the oracle properties of shrinkage estimators, leading to estimates that may be non-sparse or biased. (ii) There is no consistent criterion for selecting tuning parameters, and the computational cost of tuning is often prohibitively high.

We propose a two-stage distributed best subset selection algorithm to address these challenges. In Stage 1, we estimate the active set of true parameters to ensure sparse estimates. In Stage 2, we perform parameter estimation restricted to the active set, achieving an error bound that matches the $\ell_2$ minimax rate. Our method effectively identifies the true active set, ensuring sparsity and achieving sparse unbiased estimates with statistical oracle properties. Furthermore, we propose a data-driven tuning criterion and demonstrate that this approach can recover the true parameters accurately.

\subsection{Our Contributions} In this paper, we present a communication-efficient distributed algorithm that addresses the abovementioned challenges. Our main contributions are as follows: 
\begin{itemize} 
\item First, we introduce an innovative two-stage distributed framework for best subset selection. Initially, we estimate the active set using a surrogate likelihood function informed by the $\ell_0$ constraint, effectively reducing dimensionality and isolating key variables. Subsequent estimations within this active set ensure the sparsity of the results. Applying the $\ell_0$ constraint yields genuinely sparse estimates, enhancing model interpretability and significantly reducing communication costs due to the inherent sparsity. Our approach extends the CSL method \citep{Wang2017Efficient, Jordan2019CommunicationEfficient}, emphasizing fully recovering the true sparse parameters.

\item  Second, from an algorithmic perspective, we propose a novel splicing technique to address subproblems constrained by $\ell_0$ norms effectively. This strategy allows us to overcome the computational challenges typically associated with enumeration. Additionally, we introduce a Generalized Information Criterion (GIC) that adaptively selects the optimal parameter sparsity.

\item  Finally, from a theoretical standpoint, we demonstrate that our method can, with high probability, accurately identify the true sparsity pattern and the active set, avoiding both over-selection and omission of variables. We prove that our estimator possesses the oracle property, which is not present in other distributed sparse estimation methods. Moreover, our theoretical analysis confirms that the error bounds of our distributed best subset selection algorithm match the minimax $\ell_2$ rate of centralized processing \citep{Raskutti2011minimax}.

\end{itemize}

\subsection{Organization}
The organization of the article is as follows: Section \ref{sec:meth} explores distributed best subset selection methods and introduces the Generalized Information Criterion (GIC) for determining the optimal model size. Section \ref{thi_theory} examines the theoretical properties of the algorithm, including the accurate recovery of the active set and associated error bounds. Section \ref{fourth_experiments} compares algorithm performance through detailed simulations and analysis of real data. Finally, Section \ref{sec-con} concludes the paper with a discussion on future research directions, and the supplementary material contains proofs of the theoretical results.

\subsection{Notation}
In this paper, we adopt the following notational conventions: Consider $\boldsymbol{\theta} = (\theta_1, \ldots, \theta_p)^\top \in \mathbb{R}^p$. The $\ell_q$ norm of $\boldsymbol{\theta}$ is denoted by $\|\boldsymbol{\theta}\|_q = (\sum_{j=1}^p |\theta_j|^q)^{1/q}$ for $q \in [1, \infty)$.By default, $\|\cdot\|$ assumes $q=2$. Let $\mathcal{S} = \{1, \ldots, p\}$. For any subset $\cA \subseteq \mathcal{S}$, we denote its complement in $\mathcal{S}$ by $\cA^c = \mathcal{S} \setminus \cA$ and its cardinality by $|\cA|$. The support set of the vector $\boldsymbol{\theta}$ is defined as $\text{supp}(\boldsymbol{\theta}) = \{j : \theta_j \neq 0\}$. For an index set $\cA \subseteq \{1, \ldots, p\}$, the notation $\boldsymbol{\theta}_{\cA} = (\theta_j : j \in \cA) \in \mathbb{R}^{|\cA|}$ represents the subvector of $\boldsymbol{\theta}$ corresponding to the indices in $\cA$. Given a matrix $\boldsymbol{M} \in \mathbb{R}^{p \times p}$, we use $\boldsymbol{M}_{\cA \times \mathcal{B}} = (M_{ij} : i \in \cA, j \in \mathcal{B}) \in \mathbb{R}^{|\cA| \times |\mathcal{B}|}$ to denote the submatrix of $\boldsymbol{M}$ with rows indexed by $\cA$ and columns by $\mathcal{B}$. For any vector $\bv$ and a set $\cA$, $\bv_{\cA}$ is defined as the vector whose $j$th entry $(\bv_{\cA})_j$ equals $v_j$ if $j \in \cA$ and is zero otherwise. The gradient of $f(\boldsymbol{\theta})$ is denoted by $\nabla f(\boldsymbol{\theta})$. The notation $\nabla_{\cA} f(\btheta)= ((\nabla f(\btheta))_j, j\in \cA) \in \mathbb{R}^{|\cA|}$   represents the components of $\nabla f(\boldsymbol{\theta})$ indexed by $\cA$. Additionally, we employ the notation $a_n \lesssim b_n$ ($a_n \gtrsim b_n$) to indicate that $a_n$ is less than (greater than) $b_n$ up to a constant factor, and $a_n \asymp b_n$ (or $a_n = O(b_n)$) to signify that $a_n$ is of the same order as $b_n$.

\section{Methods}
\label{sec:meth}
This section presents an efficient algorithm for best subset selection in distributed systems. The algorithm addresses the scenario where datasets are stored and processed on multiple interconnected machines. First, based on the CSL function and the $l_0$ constraint, we propose a two-stage approach: recovering the active set, followed by parameter estimation. Further, we employ a splicing technique to speed up the solution process. Finally, the generalized information criterion (GIC) determines the optimal subset size.

\subsection{ Problem Setup and Algorithm Framework}

In a distributed computing environment, the comprehensive dataset of assets, denoted as $\{\bx_i, y_i\}_{i=1}^{N}$, is distributed across $m$ nodes. Here, $y_i \in \mathbb{R}$ represents the response variable, and $\bx_i$ is a $p$-dimensional predictor vector. The dataset is partitioned among one central machine and $m-1$ local machines, all interconnected. We represent the entirety of the samples with $\bX = \left(\bx_1^{\top}, \ldots, \bx_N^{\top}\right)^{\top}$ and $\bY = \left(y_1, \ldots, y_N\right)^{\top}$.
Each machine, indexed by $k=1,\ldots,m$, holds a subset of $n_k$ observations, symbolized as $(\bX_k, \bY_k )=\{(\bx_i, y_i)\}_{i\in M_k}$, where the $M_k$ sets are disjoint, collectively covering the entire index set $\{1,\ldots,N\}$, with the total number of observations $N$ being equally distributed across the machines such that $N=\sum_{k=1}^{m}|M_k|=\sum_{k=1}^{m}n_k$. For simplicity, we assume that $n_1=n_2=\ldots=n_k=n$ and that $N=n \cdot m$ is perfectly divisible by $m$.

We focus on the sparse linear model:
\begin{equation} \label{model}
\bY=\bX\btheta+ \bepsilon,
\end{equation}
where $\bepsilon$ denotes the error vector. A well-known challenge is the best-subset selection that seeks to minimize the empirical risk function subject to a cardinality constraint:
\begin{equation} 
   \min_{\btheta \in  \mathbb{R}^{p\times 1}} f(\btheta)=\frac{1}{2N}\|\bY-\bX\btheta\|_2^2, \quad \text{subject to} ~ \|\btheta\|_0\leq s, 
\end{equation}
where $\|\btheta\|_0=\sum_{i=1}^{p}I(\theta_i\neq 0)$ represents the $l_0$ norm of $\btheta$.  Let $\btheta^*$ be the true regression coefficient with the sparsity level $s^*$ in model \eqref{model}.
On the $k$th machine, a local quadratic loss function is defined as:
$$
f_k(\btheta)= \frac{1}{2n}\sum_{i \in {M_k}}(y_i-\bx_i^T\btheta)^2
= \frac{1}{2}\btheta^{\top}\bSigma_k\btheta-\bw_k^{\top} \btheta+ \frac{1}{2n}\sum_{i \in {M_k}}y_i^2,
$$
with $\bw_k=\frac{1}{n} \sum_{i \in {M_k}} \bx_i y_i$ and $\bSigma_k=\frac{1}{n} \sum_{i\in M_k}\bx_i\bx_i^{\top}$. The global vectors $\bw$ and $\bSigma$ are obtained by averaging across all machines: $\bw=\frac{1}{m} \sum_{k=1}^{m} \bw_k$ and $\bSigma=\frac{1}{m}\sum_{k=1}^m \bSigma_k$.
For the distributed best-subset selection problem, the global loss function can be expressed as:
$$
f(\btheta)=\frac{1}{2N}\|\bY-\bX\btheta \|_2^2=\frac{1}{m}\sum_{k=1}^{m}f_k(\btheta)= \frac{1}{2}\btheta^{\top}\bSigma\btheta-\bw^{\top} \btheta + \frac{1}{2N} \sum_{i=1}^N y_i^2.
$$
Inspired by pioneering work in communication-efficient algorithms, such as that by \cite{Jordan2019CommunicationEfficient}, we adopt an initial value $\btheta_0$ to define the gradient-enhanced loss function that approximates the global loss. This function is formulated as:
$$
l(\btheta) =f_1( \btheta ) +\left[ \nabla f( \btheta_0) -\nabla f_1( \btheta_0 ) \right]^{\top} \btheta =\frac{1}{2}\btheta ^{\top}\bSigma _1\btheta 
+\left[ \bSigma \btheta_0-\bSigma _1 \btheta_0-\bw \right]^{\top} \btheta.
$$
Then, omitting some constants, the estimator is derived by solving:
\begin{equation} \label{eq2.3}
    \min_{\btheta \in \mathbb{R}^p}~  l(\btheta) = \frac{1}{2}\btheta ^{\top}\bSigma _1\btheta 
+\left[ \bSigma \btheta_0-\bSigma _1 \btheta_0-\bw \right]^{\top} \btheta, \quad \text{subject to } \|\btheta\|_0 \le s.
\end{equation}

Our DBESS algorithm unfolds based on Eq. (\ref{eq2.3}) and assumes that $\|\btheta\|_0=s$, where the GIC will subsequently determine the optimal $s$. The algorithm is performed in two stages:
\begin{itemize}
    \item \textbf{Stage 1:} Solve Eq. (\ref{eq2.3})  and iterate with its result as the initial value until the active set is stabilized, using this stable set to estimate the active set of the true parameters.
    \item \textbf{Stage 2:} Once the active set is determined, parameter estimation is performed only within the active set. For example, the one-shot average method can be used, where local machines perform least squares estimation within the active set and then average the local estimators.
\end{itemize}

It is important to note that in Stage 1, we focus on the recovery of the active set rather than directly optimizing the CSL function under the $l_0$ constraint for parameter estimation. Stabilization of the active set is easier to achieve than convergence of the parameter estimation. In  Stage 2, after the active set has been recovered, various distributed algorithms mentioned in the introduction can be applied. The one-shot average method is simply one example of these implementations. The detailed algorithmic procedure is provided in Algorithm \ref{alg1}.

Compared to the CSL method by \cite{Jordan2019CommunicationEfficient} and CEASE method by \cite{Fan2021CommunicationEfficient}, which are directly based on the CSL function for parameter estimation, our two-stage DBESS approach offers significant advantages. Firstly, it exploits the properties of the high-dimensional sparse model and focuses on the accurate inscription of the true active set, which results in faster convergence and lower communication costs. The iterations required to stabilize the active set are typically fewer than those needed for parameter convergence, and it only involves the parameters on the active set, which is equivalent to reducing the computation from p-dimensional variables to s-dimensional, significantly reducing the cost of inter-machine communication, which is extremely beneficial for distributed computing. Secondly, since the CSL function only approximates the global loss, parameter estimations derived from it can exhibit considerable bias. The DBESS method can achieve a higher parameter estimation accuracy after accurately recovering the true active set.

\begin{algorithm}
\caption{Distributed Best-Subset Selection with a given support size s. (DBESS.Fix)}
\label{alg1}
\begin{algorithmic}[1]
\State \textbf{Input}:Datasets $\{(\bX_k,\bY_k)\}_{k=1}^m$, initial value $\btheta_0$, number of iterations $T$, support size $s$.
\State Initialize active set $\cA_0 = \{i \mid |(\btheta_0)_i| \text{ is among the } s \text{ largest in } \btheta_0\}$, $t=0$.
\While{$t<T$}
\For{$k=1,2,...,m$}
\State On machine $k$, compute the gradient $\nabla f_k(\btheta_t) = \mathbf{\Sigma}_k \btheta_t - \mathbf{w}_k$ and broadcast it to machine 1.
\EndFor
\State On machine 1, aggregate gradients and solve for $\btheta_{t+1}$:
\begin{equation} \label{eq2.1:theta}
\btheta_{t+1} 
        = \argmin_{\| \btheta\|_0 = s}\frac{1}{2}\btheta ^{\top}\bSigma _1\btheta 
        +\left[ \bSigma \btheta _t-\bSigma _1\btheta _t-\bw \right]^{\top} \btheta.
\end{equation}
\State Update active set $\cA_{t+1}=\left\{j: (\btheta_{t+1})_j \neq 0\right\}$. 
\If{$\cA_{t+1} = \cA_{t}$}
        \State \textbf{break}
\EndIf
\State $t=t+1$.
\EndWhile
\State Machine 1 broadcasts  active set $\hat \cA =\cA_{t+1}$ to all machines.
\For{$k=1,2,...,m$}
\State On machine $k$, perform least squares estimation restricted to $\hat \cA$:
$$\hat{\btheta}^k = \argmin_{\btheta_{(\hat \cA)^c}=\boldsymbol{0}} \|\bY_k - \bX_k \btheta\|^2.$$
\State Send $\hat{\btheta}^k$ to machine 1.
\EndFor
\State On machine 1, compute the average: $\hat\btheta = \frac{1}{m} \sum_{k=1}^m \hat{\btheta}^k$.

\State \textbf{Output}: $\hat\btheta,~\hat \cA$.
\end{algorithmic}
\end{algorithm}

\textbf{Remark 2.1:}  Lines 3-13 of Algorithm \ref{alg1} are dedicated to estimating the active set via a CSL function. Subsequently, lines 14-19 describe the process where local machines conduct least squares estimation constrained to the active set, culminating in an aggregation step on machine 1.

\textbf{Remark 2.2:} When the initial value $\btheta_0$ is close to the true parameter $\btheta^*$, Algorithm \ref{alg1} can recover the true active set immediately after one iteration. Therefore, for the theoretical analysis in Section \ref{thi_theory}, we will set assumptions on the initial values to show that the one-step DBESS estimator is statistically efficient.

\newpage

\subsection{Solving Sub-problems with Splicing Techniques}
In this subsection, we will employ the splicing technique to address the subproblem defined by (\ref{eq2.1:theta}). Without loss of generality, we examine a typical quadratic function as follows:
\begin{equation} \label{eq2.5}
     \min_{\btheta} \quad l_n(\btheta)=\frac{1}{2}\btheta^{\top} \bG \btheta +\bb^{\top} \btheta, \quad \text{subject to } \|\btheta\|_0=s.
\end{equation}

It becomes evident that in the context of the subproblem (\ref{eq2.1:theta}), we have $\bG=\bSigma_1, \bb=\bSigma \btheta_0-\bSigma _1 \btheta_0-\bw$.
Then, the augmented Lagrangian form  of problem \eqref{eq2.5} is
\begin{equation} \label{2.2.2}
\begin{aligned}
    \min_{\btheta, \bd, \bv} \quad  &L_\rho (\btheta, \bv, \bd) 
    = \frac{1}{2} \btheta^{\top} \bG\btheta + \bb^{\top}\btheta + \bd^{\top}(\btheta-\bv) 
     + \frac{\rho}{2}\|\btheta - \bv \|^2, \\
    \text{subject to } &\|\bv\|_0=s.
\end{aligned}
\end{equation}
where $\rho$ is a hyperparameter. We can derive the optimal conditions for the problem as follows:
\begin{lemma} \label{lem2.1}
     Suppose $\left(\btheta^{\diamond}, \boldsymbol{v}^{\diamond}, \boldsymbol{d}^{\diamond}\right)$ is a coordinate-wise minimizer of (\ref{2.2.2}). Denote $\mathcal{A}^{\diamond}=\{j \in$ $\left.\mathcal{S}: \boldsymbol{v}_j^{\diamond} \neq 0\right\}$ and $\mathcal{I}^{\diamond}=\left(\mathcal{A}^{\diamond}\right)^c$. Then $\left(\boldsymbol{v}^{\diamond}, \boldsymbol{\beta}^{\diamond}, \boldsymbol{d}^{\diamond}\right)$ and $\left(\mathcal{A}^{\diamond}, \mathcal{I}^{\diamond}\right)$ satisfy:
$$
\begin{aligned}
& \mA^{\diamond}=\left\{j: \sum_{i=1}^p \mI\left(\left|\btheta_j+\frac{1}{\rho} \bd_j\right| \leqslant\left|\btheta_i+\frac{1}{\rho} \bd_i\right|\right) \leqslant s\right\}, \quad\mI^{\diamond}=(\mA^{\diamond})^c, \\
&\btheta^{\diamond}_{\mA^{\diamond}}=-\left(\bG_{\mA^{\diamond}\times \mA^{\diamond}}\right)^{-1}\bb_{\mA^{\diamond}}, \quad  \btheta^{\diamond}_{\mI^{\diamond}}=0,\\
& \bd^{\diamond}_{\mA^{\diamond}}=0, \quad \bd^{\diamond}_{\mI^{\diamond}}=-(\bG_{\mI^{\diamond}\times \mI^{\diamond}}) \btheta^{\diamond}_{\mI^{\diamond}}-\bb_{\mI^{\diamond}},\\
& \bv^{\diamond}=\btheta^{\diamond}.
\end{aligned}
$$
\end{lemma}
According to lemma \ref{lem2.1}, assume \((\mA^q,\mI^q,\btheta^q,\bd^q)\) is the solution at the \(q\)-th iteration, the active set for the \(q+1\)-th iteration updates as:
\begin{equation} \label{2.2.3}
    \mA^{q+1}=\left\{j: \sum_{i=1}^p I\left(\left|\btheta^q_j+\frac{1}{\rho} \bd^q_j\right| \leqslant\left|\btheta^q_i+\frac{1}{\rho} \bd^q_i\right|\right) \leqslant s\right\}, \quad \mI^{q+1}=(\mA^{q+1})^c.
\end{equation}
Then, the updates for the primal and dual vectors are given by:
\[
\begin{aligned}
\btheta^{q+1}_{\mA^{q+1}} & =-\left(\bG_{\mA^{q+1}\times \mA^{q+1}}\right)^{-1}\bb_{\mA^{q+1}}, \quad\btheta^{q+1}_{\mI^{q+1}}=0, \\
\bd^{q+1}_{\mA^{q+1}} & =0, \quad \bd^{q+1}_{I^{q+1}}=-(\bG_{\mI^{q+1}\times \mI^{q+1}}) \btheta^{q+1}_{\mI^{q+1}}-\bb_{\mI^{q+1}}.
\end{aligned}
\]
The penalty parameter \(\rho\) plays a crucial role in the update of the active set, as seen in equation (\ref{2.2.3}). A larger \(\rho\) results in more significant updates to the active set, while a smaller \(\rho\) results in minor updates. From another perspective, \(\rho\) can be related to the number of exchanges \(C\) between the active and inactive sets. We can formally define:

\[
\begin{aligned}
\mS_{C, 1}^q & = \left\{j \in \mA^q: \sum_{i \in \mA^q} \mI\left( |\btheta_j^q + \frac{1}{\rho} \bd_j^q| \geq |\btheta_i^q + \frac{1}{\rho} \bd_i^q| \right) \leq C \right\} \\
& = \left\{j \in \mA^q: \sum_{i \in \mA^q} \mI\left( |\btheta_j^q| \geq |\btheta_i^q| \right) \leq C \right\},
\end{aligned}
\]

\[
\begin{aligned}
\mS_{C, 2}^q & = \left\{j \in \mI^q: \sum_{i \in \mI^q} I\left( |\btheta_j^q + \frac{1}{\rho} \bd_j^q| \leq |\btheta_i^q + \frac{1}{\rho} \bd_i^q|\right) \leq C \right\} \\
& = \left\{j \in \mI^q: \sum_{i \in \mI^q} I\left( |\bd_j^q| \leq |\bd_i^q| \right) \leq C \right\}.
\end{aligned}
\]
Indeed, \(\mS_{C,1}^q\) and \(\mS_{C,2}^q\) can be discerned as representing the least significant \(C\) variables within \(\mA^q\) and the most pivotal \(C\) variables within \(\mI^q\), respectively. To elaborate further, we introduce the following lemma:
\begin{lemma}{}
\begin{enumerate}
    \item  \textbf{Backward sacrifice:} For any $j\in \mathcal{A}$, the magnitude of removing the $j$-th variable from $\mathcal{A}$ is
        \[
        \xi_j^*=l_n\left(\left.\hat{\btheta}\right|_{\mathcal{A} \backslash\{j\}}\right)-l_n(\hat{\btheta})= \frac{1}{2}G_{jj}(\hat\btheta_j)^2.
        \]
    \item \textbf{Forward sacrifice:} For any $j\in \mathcal{I}$, adding the $j^{th}$ variable to $\mathcal{A}$ is
        \[
        \zeta_j^*=l_n(\hat{\btheta})-l_n\left(\hat{\btheta}+\left.\hat{t}\right|_j\right)=\frac{1}{2} \left[G_{jj}\right]^{-1} \hat d_j^2,
        \]
        where $\left.\hat{t}\right|_j=\argmin_{t}l_n(\hat{\btheta}+t|_j).$ 

\end{enumerate}
  
\end{lemma}
Drawing from previous discussions, it becomes clear that as \(C\) increases, the number of elements exchanged between the active and inactive sets during each iteration also increases. This suggests that fewer iterations are needed to achieve convergence. Therefore, \(C\) can be viewed as a trade-off between the computational workload of each iteration and the overall number of iterations required. This role of \(C\) is reminiscent of the function of the penalty parameter \(\rho\) mentioned earlier. We have a lemma that further delineates this relationship in finer detail.

\begin{lemma}
Let $C \) denote the size of the exchanged subset of elements. For any positive integer $C \) such that $C \leq |\mathcal{A}^q| \), the associated range for $\rho \) during the $q\)-th iteration is as follows:
    $$
    \rho \in
\begin{cases}
\left( \frac{\min_{j \in \mS_{C,2}^q} |d_j^q|}{\max_{i \in \mS_{C,1}^q} |\theta_i^q|} , +\infty\right) &  \text{for } C=0,\\

\left( \frac{\min_{j \in \mS_{C+1,2}^q} |d_j^q|}{\max_{i \in \mS_{C+1,1}^q} |\theta_i^q|} , \frac{\min_{j \in \mS_{C,2}^q} |d_j^q|}{\max_{i \in \mS_{C,1}^q} |\theta_i^q|}) \right]
&  \text{for } 1\le C <s, \\

\left(0,\frac{\min_{j \in \mS_{C,2}^q} |d_j^q|}{\max_{i \in \mS_{C,1}^q} |\theta_i^q|} \right] 
& \text{for } C=s. \\

\end{cases}
$$
\end{lemma}

This lemma elucidates the interplay between the penalty parameter $\rho$ in the augmented Lagrange function and the exchange counts $C$ between active and inactive sets. Specifically, while $C$ dictates the frequency of exchanges between these sets, convergence is ensured for any value of $C$, given the finite combinations $C_n^s$. Determining an optimal $C$ is more straightforward than fine-tuning $ \rho $. A judicious approach for loss minimization in problem (\ref{eq2.1:theta}) involves choosing a $C$ that results in a subsequent reduction in loss after set updates. This procedure is denoted as ``Splicing". By incorporating splicing throughout iterations, we introduce the Splicing algorithm, detailed in Algorithm \ref{alg3:Quad_Splicing}, which adeptly tackles the subproblem (\ref{eq2.1:theta}).

\begin{algorithm}[H]
\caption{Splicing Algorithm for Quadratic Loss Function(Quad\_Splicing)}
\label{alg3:Quad_Splicing}
\begin{algorithmic}[1]
\State \textbf{Input:} $\bG, \bb$, initial active set $\mathcal{A}^0$ with $s$ elements, maximum number of exchange $C_{\max}\le s$, threshold $\tau_s$.
\State \textbf{Initialize:} $q =-1$, $ \mathcal{I}^0 =\left(\mathcal{A}^0\right)^c,$
 $\btheta^0 =\argmin _{\btheta_{\mathcal{I}^0=0}} \frac{1}{2}\btheta ^{\top} \bG\btheta +\bb^{\top} \btheta,$
and $\bd^0=-\bG \btheta^0 -\bb$.
\Repeat
    \State $q =q+1$,  $L = l_n(\btheta^q)$.
    \For{$C=1, \ldots, C_{\max }$}
       \State Compute $\xi_j =G_{jj}(\theta_j^q)^2, j \in \mathcal{A}^q$,  and $\zeta_j=G_{jj} ^{-1} (d_j^q)^2, j \in \mathcal{I}^q$.
       \State Update active set: $\tilde{\mathcal{A}} =(\mathcal{A}^q \backslash \mathcal{S}_{C, 1}^q) \cup \mathcal{S}_{C, 2}^q$, where
            \begin{align*}
            \mathcal{S}_{C, 1}^q=&\left\{j \in \mathcal{A}^q: \sum_{i \in \mathcal{A}^q} \mathrm{I} (\xi_j \geq \xi_i ) \leq C\right\}, \\
            \mathcal{S}_{C, 2}^q=&\left\{j \in \mathcal{I}^q: \sum_{i \in \mathcal{I}^q} \mathrm{I} (\zeta_j \leq \zeta_i ) \leq C\right\}.
            \end{align*}
       \State Set $\tilde{\mathcal{I}} \leftarrow(\tilde{\mathcal{A}})^c$,  update $\tilde{\btheta} = \argmin_{\btheta_{\tilde{\mathcal{I}}}=0} l_n(\btheta)$.
       \State Compute $\tilde{\bd} =-\bG \tilde\btheta-\bb $.
       \If{$L-l_n(\tilde{\btheta})>\tau_s$}
           \State $ L=l_n(\tilde{\btheta}),$
           $(\mathcal{A}^{q+1}, \mathcal{I}^{q+1}, \btheta^{q+1}, \bd^{q+1}) =(\tilde{\mathcal{A}}, \tilde{\mathcal{I}}, \tilde\btheta, \tilde{\bd})$,
           and \textbf{break}.
       \EndIf
    \EndFor
\Until{$\mathcal{A}^{q+1}=\mathcal{A}^q$}
\State \textbf{Output:} $(\btheta^q, \mathcal{A}^q)$.
\end{algorithmic}
\end{algorithm}
\noindent\textbf{Remark 2.3:} Selecting an appropriate initial active set, \(\mathcal{A}_0\), can significantly speed up the algorithm's convergence. The closer $\mathcal{A}_0$ is to the true active set, the fewer exchange iterations the splicing algorithm has to perform, resulting in quicker convergence. Moreover, even with a less-than-ideal choice of \(\mathcal{A}_0\), the increase in the number of exchange iterations required would be minimal.

\noindent\textbf{Remark 2.4:}
A threshold $\tau_s > 0$ can enhance the algorithm's efficiency, facilitating faster convergence and reducing redundant iterations.

\noindent\textbf{Remark 2.5:}
The parameter \(C_{\text{max}}\), a positive integer no greater than \(s\), defines the maximum number of exchanges allowed in Algorithm \ref{alg1}. Empirical evidence suggests that setting \(C_{\text{max}}\) to either 2 or 5 frequently leads to enhanced efficiency of the algorithm.

\subsection{GIC for Optimal Subset Size Selection}
Continuing from Algorithm \ref{alg1}, the next step in the method revolves around recovering the true sparsity \(s^*\). Model selection techniques such as cross-validation and information criteria are commonly employed. In this section, we introduce a data-driven generalized information criterion (GIC) tailored to select the optimal sparsity adaptively. Specifically, the GIC is defined as:
$$
\text{GIC}(\btheta)= N \log \|\bY-\bX \btheta \|_2^2 +  \|\btheta\|_0 \log p \log \log N.
$$

The GIC encompasses two components: the first term is the loss function, while the second acts as a penalty to curb model complexity, thereby preventing overfitting. To implement this criterion, execute Algorithm \ref{alg1} for each \(s=1,2,.., s_{\max}\). Subsequently, select the sparsity level \(s\) that minimizes the GIC. This procedure culminates in the complete adaptive Distributed Best-Subset Selection  Algorithm. The complete details of the algorithm are presented in Algorithm \ref{alg5:dbess}. Moreover, Theorem 3.2 establishes that with such a GIC, the true sparsity $s^* \) can be accurately recovered.

\begin{algorithm}
\caption{Distributed Best-Subset Selection(DBESS)
}
\label{alg5:dbess}
\begin{algorithmic}[1]
\State \textbf{Input}: Datasets $\{(\bX_k,\bY_k)\}_{k=1}^m$, initial value $\btheta_0$, maximum sparsity $s_{\max}$.
\For{$s=1,2,...,s_{\max}$}
    \State $(\hat\btheta^s, \hat \cA^s)=$ DBESS.Fix$(\{(\bX_k,\bY_k)\}_{k=1}^m,\btheta_0, s).$
    \State Computing GIC: $\text{GIC}(s)=\text{GIC}(\hat\btheta^s).$
\EndFor
\State $s_{\min}=\argmin_{s} \text{GIC}(s).$
\State \textbf{Output}: $\hat\btheta^{s_{\min}},\hat\cA^{s_{\min}}$.
\end{algorithmic}
\end{algorithm}

\section{Theoretical Properties}
\label{thi_theory}
This section delves into our algorithm's statistical performance guarantees and convergence analysis, as detailed in Section \ref{subsec-pro}. Before outlining these guarantees formally, we will explore the requisite technical conditions in Section \ref{subsec-ass}. The details of the proof can be found in the supplementary material.

\subsection{Assumptions}
\label{subsec-ass}
Here, we introduce the common assumptions in best subset selection and distributed statistical inference, which serve as a foundation for subsequent theoretical analysis.

\begin{enumerate}[label=(A\arabic*),leftmargin=*]

    \item \textbf{Sub-Gaussian Residual:} 
    The random errors $\epsilon_1, \ldots, \epsilon_N$ are i.i.d with mean zero and sub-Gaussian tails, i.e. there exists a $\sigma>0$ such that $P\left\{\left|\epsilon_i\right| \geq\right.$ $t\} \leq 2 \exp \left(-t^2 / \sigma^2\right)$, for all $t \geq 0$.

    \item \textbf{Spectral Restricted Condition (SRC):}
    For every $k = 1, 2, \dots, m$ and any set $\mathcal{A}$ with $|\mathcal{A}| \le s$, we have
    \[
    m_s \le \lambda_{\text{min}}(\bSigma_{k, \mathcal{A} \times \mathcal{A}}) \le \lambda_{\text{max}} (\bSigma_{k, \mathcal{A}\times \mathcal{A}}) \le M_s,
    \]
    where \(\lambda_{\text{min}}(A)\) and \(\lambda_{\text{max}}(A)\) represent the smallest and largest eigenvalues of the matrix \(A\), respectively.
    \item \textbf{Characterization of Non-diagonal Elements:}
    For every $k = 1, 2, \dots, m$ and any sets $\mathcal{A}$ and $\mathcal{B}$ satisfying $|\mathcal{A}| \leq s$, $|\mathcal{B}| \leq s$, and $\mathcal{A} \cap \mathcal{B} = \emptyset$, we have
    \[
    \lambda_{\text{max}}\left(\bSigma_{k,\mathcal{A}\times \mathcal{B}}\right) \leq \nu_s.
    \]
    
    Define constants $C_1$ and $C_2$ as
    \begin{align*}
    C_1 &:= \left(\frac{M_s}{2} + \frac{\nu_s^2}{m_s} + \frac{M_s \nu_s^2}{2 m_s^2}\right)\left[\left(2+\Delta+\frac{2 \nu_s}{m_s}\right)^2+(2+\Delta)^2\right]\left(\frac{\nu_s}{m_s}\right)^2, \\
    C_2 &:= \frac{m_s}{2} - \frac{\nu_s^2}{m_s} - \frac{M_s \nu_s^2}{2 m_s^2},
    \end{align*}
    and there exists a constant $\Delta > 0$ such that $\rho_s := \frac{C_1}{(1-\Delta) C_2} < 1$.

    \item \textbf{Minimum Signal Condition :} 
     Let $b^* = \min_{j \in A^*} |\theta_j|^2$, then $\frac{1}{b^*} = o\left(\frac{N}{s \log p \log \log N}\right)$.

    \item \textbf{Sparsity Condition:} 
     $\frac{s^* \log p \log (\log N)}{N} = o(1)$ and $\frac{s_{\max} \log p}{N} = o(1)$.

    \item \textbf{Threshold Condition:}
    The threshold $\tau_s$ satisfies $\tau_s = O\left(\frac{s \log p \log \log N}{N}\right)$.

    \item \textbf{Initial Condition:}
    Given $\bdelta =\btheta_0 - \btheta^*$ with $\btheta_0$ as the initial value, then $\|(\bSigma-\bSigma_1)\bdelta\|_2 \lesssim \sqrt{\frac{s \log p}{N}}$.

\end{enumerate}

The sub-Gaussian assumption (A1) is a standard premise frequently found in related literature. Building on \cite{Huang2018Constructive} and \cite{Zhu2020Polynomial}, assumption (A2) introduces the Spectral Restricted Condition (SRC), which sets spectral limits for the diagonal submatrices of the covariance matrix. In distributed settings, we assume that each of the $m$ machines adheres to these spectral boundaries, ensuring a homogeneous and consistent environment across the distributed framework. Assumption (A3) imposes restrictions on spectral bounds, aligning with Assumption 3 in \cite{Zhu2020Polynomial}. The characterization of the strength of minimum signals (A4) echoes the conditions necessary for consistent variable selection in high-dimensional contexts, as outlined by \cite{Fan2013Tuning} and \cite{Zhu2020Polynomial}. Assumption (A5) ensures a manageable relationship between sparsity level and dimensionality, vital for subset selection consistency under the GIC. The threshold $\tau_s$ in (A6) effectively controls random errors, reducing unnecessary iterations in the algorithm. In other words,  the condition (A6) is primarily an algorithmic necessity rather than a statistical prerequisite. Assumption (A7) constrains the initial values, aiming to ensure that the active set identified by the one-step DBESS estimator aligns with the true active set and that the estimator's error attains the same minimax error rate as centralized processing. This condition is actually quite relaxed: considering that $\|(\bSigma-\bSigma_1)\|_2 = O\left(\sqrt{\frac{\log p}{n}}\right)$, and selecting $\btheta_0$ as the lasso estimator from machine 1, which satisfies $\|\btheta_0 - \btheta^* \|_2 \lesssim \sqrt{\frac{s\log p}{n}}$, it only requires that $n \gtrsim m \log p$ to satisfy condition (A7). These conditions are either comparable to or less restrictive than those in \citet{Battey2018} and \citet{Jordan2019CommunicationEfficient}.

\subsection{Statistical Guarantees}
\label{subsec-pro}
First, we demonstrate that the DBESS method can effectively recover the true active set, identifying the true effective variables with a high probability.
\begin{theorem}
When $s^* \leq s $, let $(\hat{\btheta}, \hat \cA)$ be the solution of Algorithm 1. Under Assumptions (A1)-(A7), we have:
$$
\mathbb{P}\left( \mathcal{A}^*  \subseteq  \hat{\mathcal{A}} \right) \geq 1-\delta_1-\delta_2-\delta_3,
$$
where $\delta_i=O\left(\exp \left\{\log p-K_i \frac{N}{s} \min _{j \in \mathcal{A}^*}\left|{\theta}_j^*\right|^2\right\}\right)$.
Asymptotically,
$$
\lim _{N \rightarrow \infty} \mathbb{P}\left(\mathcal{A}^* \subseteq \hat{\mathcal{A}}\right)=1 .
$$
Especially, if $s=s^{\star}$, then we have
$$
\lim _{N \rightarrow \infty} \mathbb{P}\left(\mathcal{A}^*=\hat{\mathcal{A}}\right)=1.
$$
\end{theorem}

The principle behind the DBESS method's ability to recover the true active set is based on the following logic. If the estimated active set misses some elements of the true active set, then through the Splicing process, the loss calculated based on the updated active set will be lower than the loss associated with the original active set. The Splicing process will continue until the estimated active set matches the true active set. At this point, the loss reaches its minimum, and the Splicing program will terminate.

Theorem 3.1 establishes that our algorithm selects the active set with high probability, thereby guaranteeing the inclusion of all essential variables without exception. This foundational result paves the way for the accurate recovery of the true active set, as detailed in Theorem 3.2.

\begin{theorem}
Assume that Conditions (A1)-(A7) hold with $s_{\max}$. Denote $(\hat\btheta^{s_{\min}},\hat\cA^{s_{\min}}) $ as the solution of Algorithm 3. Under the  GIC, with probability $1-O\left(p^{-\alpha}\right)$, for some positive constant $\alpha>0$ and a sufficiently large $N$, the algorithm selects the true active set, that is, $\hat{\mathcal{A}}^{s_{\min }}=\mathcal{A}^*$.
\end{theorem}

Theorem 3.2 indicates that even if the true sparsity of the model is unknown, the DBESS method based on the GIC can recover the true sparsity with high probability. Furthermore, in conjunction with Theorem 3.1, the DBESS algorithm can identify relevant variables with extremely high accuracy without including extraneous variables or omitting any pertinent ones. After accurately identifying the true active set, we can focus on estimating the parameters within this set. At this point, various distributed computing methods can be applied, such as a divide-and-conquer strategy for estimation.
It is important to note that because the elements outside the active set are constrained to zero, only the parameters within the active set are effective. This means that the sparsity of the parameters is maintained even after the averaging process, thus preserving the interpretability of the sparse model.

Next, we demonstrate that the estimation error obtained by the DBESS method achieves the minimax rate. In the context of high-dimensional sparse linear regression, consider the $\ell_0$-ball $\mathcal{B}\left(s^*\right)$ defined as
$$
\mathcal{B}\left(s^*\right) = \{\boldsymbol{\theta} \in \mathbb{R}^p : \|\boldsymbol{\theta}\|_0 \leq s^*\}.
$$
\cite{Raskutti2011minimax} established a well-known minimax lower bound for the $\ell_2$-norm of estimators. For any estimator $\widehat{\boldsymbol{\theta}}_N$ based on $N$ samples, the following bound applies:
$$
\min_{\hat{\boldsymbol{\theta}}_N} \max_{\boldsymbol{\theta}^* \in \mathcal{B}\left(s^*\right)} \mathbb{E}_{\btheta}\left\|\hat{\boldsymbol{\theta}}_N - \boldsymbol{\theta}^*\right\|_2^2 = O\left(\frac{\sigma^2 s^* \log(p / s^*)}{N}\right).
$$
For the estimates $\hat\btheta$ obtained via the DBESS method, the following theorem is presented:

\begin{theorem}
 Let $\hat \btheta$ be the coefficient estimator outputted by Algorithm 3. Suppose Assumptions in Theorem 3.1 hold, then the following bound hold
$$
\mathbb{E}\left[\left\|\hat{\btheta}-\btheta^*\right\|_2^2\right] 
 \leq  C_6 \frac{\sigma^2s^* \log \left(p / s^*\right)}{N},
$$
where $C_6$ is a constant independent of $(n,m,p,s^*)$.
\end{theorem}

Theorem 3.3 demonstrates that the estimation error achieved by the DBESS method reaches the optimal minimax rate within a distributed setting, aligning with the rates found in centralized frameworks. Compared with Theorem 6 from  \cite{Jordan2019CommunicationEfficient}, the error boundary for the one-step CSL method is depicted as $O\left(\left(\sqrt{\frac{s^* \log p}{N}}+ \frac{(s^*)^{3 / 2} \log p}{n}\right)^2\right)$. The local sample size $n$ is required to meet the condition $n \gtrsim m(s^*)^2\log p$ to attain optimal minimax rate. However, the DBESS algorithm relaxes this stipulation to $n \gtrsim m\log p$. This relaxation enhances the method's practical applicability, an advantage  that will be demonstrated in subsequent numerical experiments.  

Because our approach accurately identifies the true active set, we can straightforwardly derive the asymptotic distribution of $\btheta$.

\begin{corollary} (Asymptotic Properties)
Assuming the assumptions of Theorem 3.3 hold, the solution $\hat{\boldsymbol{\theta}}^{s_{\min}}$ of DBESS is, with high probability, the oracle estimator. Specifically,
$$
P\left\{\hat{\boldsymbol{\theta}}^{s_{\min}} = \hat{\boldsymbol{\theta}}^o\right\} = 1 - O\left(p^{-\alpha}\right),
$$
where $\alpha > 0$ and $\hat{\boldsymbol{\theta}}^o$ denotes the least squares estimate obtained by aggregating the data on a single machine, given the true active set $\mathcal{A}^*$. Furthermore,
$$
\hat{\boldsymbol{\theta}}^{s_{\min}}_{\mathcal{A}^*} \sim N\left(\boldsymbol{\theta}_{\mathcal{A}^*}^*, \boldsymbol{\Sigma^*}\right),
$$
where $\boldsymbol{\Sigma^*} = \left(\boldsymbol{X}_{\mathcal{A}^*}^{\top} \boldsymbol{X}_{\mathcal{A}^*}\right)^{-1}$.
\end{corollary}
Corollary 3.4 demonstrates that our DBESS estimator possesses the oracle property, a feature absent in previous distributed sparse estimation methods that rely on approximate functions or $\ell_1$ penalty.

\section{Numerical Experiments}
\label{fourth_experiments}
In this section, we comprehensively evaluate the proposed algorithm's optimization and statistical performances, juxtaposing it with prevailing distributed sparse algorithms in the literature. The organization of this section is structured as follows to aid the reader's comprehension. Section 4.1 delineates the methodologies for generating simulated data, enumerates the distributed algorithms selected for comparative analysis, and outlines the metrics adopted for evaluating performance. Section 4.2 delves into comparing estimation errors across algorithms as a function of iteration numbers, thereby elucidating their respective convergence rates and error properties. Section 4.3 assesses the algorithms' efficacy in accurately identifying the true active set, featuring an analysis of critical performance indicators such as the True Positive Rate (TPR) and True Negative Rate (TNR). Section 4.4 investigates the impact of varying the number of machines and local sample sizes on algorithmic performance.

\subsection{Experimental Setup}
\textbf{Data Generation:} This study constructs a suite of synthetic datasets to rigorously evaluate the algorithm's performance across various problem dimensions and configurations. A multivariate Gaussian data matrix \(\bX\) is synthesized following a multivariate normal distribution \(\bX \sim \mathrm{MVN}(\boldsymbol{0}, \bSigma)\). A sparse coefficient vector \(\btheta^*\) with \(s^*\) non-zero elements, where half of the non-zero entries are set to $1$ and the other half to $-1$. The response vector \(\bY\) is then generated using \(\bY = \bX\btheta^* + \bepsilon\), where the error term \(\epsilon_i\) follows an independent and identically distributed normal distribution \(\epsilon_i \sim N(0,\sigma^2)\), independent from \(\bX\). The noise level \(\sigma\) is set to achieve a signal-to-noise ratio (SNR) of $1$, defined as \(\text{SNR} = \frac{\text{Var}(\bX \btheta^*)}{\text{Var}(\bepsilon)} = \frac{(\btheta^*)^\top \bSigma \btheta^*}{\sigma^2}\). Finally, the data pair \((\bX, \bY)\) is evenly partitioned into \(m\) segments and distributed among \(m\) computational units to simulate a distributed computing environment.

We define the covariance matrix \(\bSigma = (\sigma_{ij})_{p \times p}\) based on established practices in distributed computing and optimal subset selection algorithms, incorporating two feature  structures for our experimental design:

\textbf{1. Uncorrelated Structure:} \(\bSigma = \operatorname{diag}(10, 5, 2, 1, \ldots, 1)\), as suggested by \cite{Fan2021CommunicationEfficient}.

\textbf{2. Correlated Structure:} \(\sigma_{ij} = 0.8^{|i-j|}\), indicating a strong correlation between features that diminishes with increasing index distance.

\textbf{Competing Algorithms and Parameter Tuning:} We evaluate the following methodologies for constructing sparse models:

\begin{itemize}
    \item \textbf{DBESS:} Our novel approach is introduced in this paper.
    
    \item \textbf{CSL:} Following \cite{Jordan2019CommunicationEfficient}, CSL utilizes a gradient-enhanced function with an $l_1$ penalty. The regularization parameter $\lambda$ is selected based on cross-validation, employing an iterative method.
    
    \item \textbf{CEASE:} Extending CSL, described by \cite{Fan2021CommunicationEfficient},  the CEASE method iteratively solves the problem using the PPA algorithm, which is based on the gradient-enhanced function and $l_1$ penalty. The PPA proximal term parameter is set to $\alpha = 0.15 \frac{p}{n}$.
    
    \item \textbf{Global Lasso:} Consolidates data on a single machine to perform variable selection using Lasso, with $\lambda$ selected based on cross-validation.
\end{itemize}

\textbf{Evaluation Metrics:} We assess subset selection performance through metrics including the True Positive Rate (TPR), True Negative Rate (TNR), and the Matthews Correlation Coefficient (MCC), alongside evaluating parameter estimation precision via Squared Estimation Error(SEE) and the Relative Error in Estimation (ReEE). Detailed explanations of these metrics and their importance are provided below:

\begin{itemize}
    \item $\operatorname{TPR} = \frac{\operatorname{TP}}{\operatorname{TP} + \operatorname{FN}},~~$
  $\operatorname{TNR} = \frac{\operatorname{TN}}{\operatorname{FP} + \operatorname{TN}},~~ \mathrm{SEE}=\|\hat{\boldsymbol{\theta}} - \boldsymbol{\theta}^*\|_2,~~$ $\operatorname{ReEE} = \frac{\|\hat{\boldsymbol{\theta}} - \boldsymbol{\theta}^*\|_2}{\|\boldsymbol{\theta}^*\|_2}$
    \item $\operatorname{MCC} = \frac{\operatorname{TP} \times \operatorname{TN} - \operatorname{FP} \times \operatorname{FN}}{\sqrt{(\operatorname{TP}+\operatorname{FP})(\operatorname{TP}+\operatorname{FN})(\operatorname{TN}+\operatorname{FP})(\operatorname{TN}+\operatorname{FN})}},$
\end{itemize}
where
$
\mathrm{TP}=\left|\widehat{\cA} \cap \cA^*\right|, \mathrm{TN}=\left|\widehat{\cI} \cap \cI^*\right|, 
\mathrm{FN}=\left|\widehat{\cI} \cap \cA^*\right|, \mathrm{FP}=\left|\widehat{\cA} \cap \cI^*\right|.
$

\subsection{ Analysis of Convergence Rates and Estimation Errors }
In this section, we assess the efficacy of different algorithms in a distributed setting, focusing on two key metrics: convergence rate and estimation accuracy. We specifically examine the required number of iterations for convergence and the estimation error at the point of convergence, where relevant.

Following the setup specified in \cite{Fan2021CommunicationEfficient}, we configure our covariance matrix as \(\boldsymbol{\Sigma} = \operatorname{diag}(10, 5, 2, 1, \ldots, 1)\) and generate simulation data following the procedures detailed in Section 4.1. We set a total sample size of \(N=10000\), with the number of machines \(m\) varying among \{10, 20, 40\}, feature dimension \(p=100\), and sparsity level \(s^*=5\). The maximum iteration limit is designated as \(\text{max\_iter} = 10\), to evaluate the error \(\| \btheta_t-\btheta^*\|_2\) across different algorithms as a function of iteration count. Additionally, we explore two initialization strategies: zero initialization $\boldsymbol{0}$ and oneshot initialization $\btheta_{\mathrm{oneshot}}$, where the latter averages least squares estimates across all machines for initial values.

Figure \ref{f1} clearly illustrates our DBESS method's superiority in error reduction and convergence speed compared to CSL and CEASE. The curves represent the average values over 100 independent runs, and the error bands correspond to one standard deviation.
This comparison showcases the effectiveness of our method DBESS relative to the CSL and CEASE methods. Additionally, we include the Global Lasso method, which consolidates all data onto a single machine and applies the lasso algorithm, optimizing the penalty parameter through cross-validation.
The first row of subplots focuses on zero initialization (\(\btheta_0= \boldsymbol{0}\)), maintaining a consistent sample size of \(N=10000\) across varying machine counts \(m\). The subplots are dedicated to distinct configurations of \((n, m)\): \((2000, 5)\), \((1000, 10)\), and \((250, 40)\). Correspondingly, the second row highlights the oneshot initialization strategy (\(\btheta_0= \boldsymbol{\theta}_{\text{oneshot}}\)).
\begin{figure}[ht]
    \centering
    \begin{subfigure}[b]{\linewidth}
        \includegraphics[width=\linewidth]{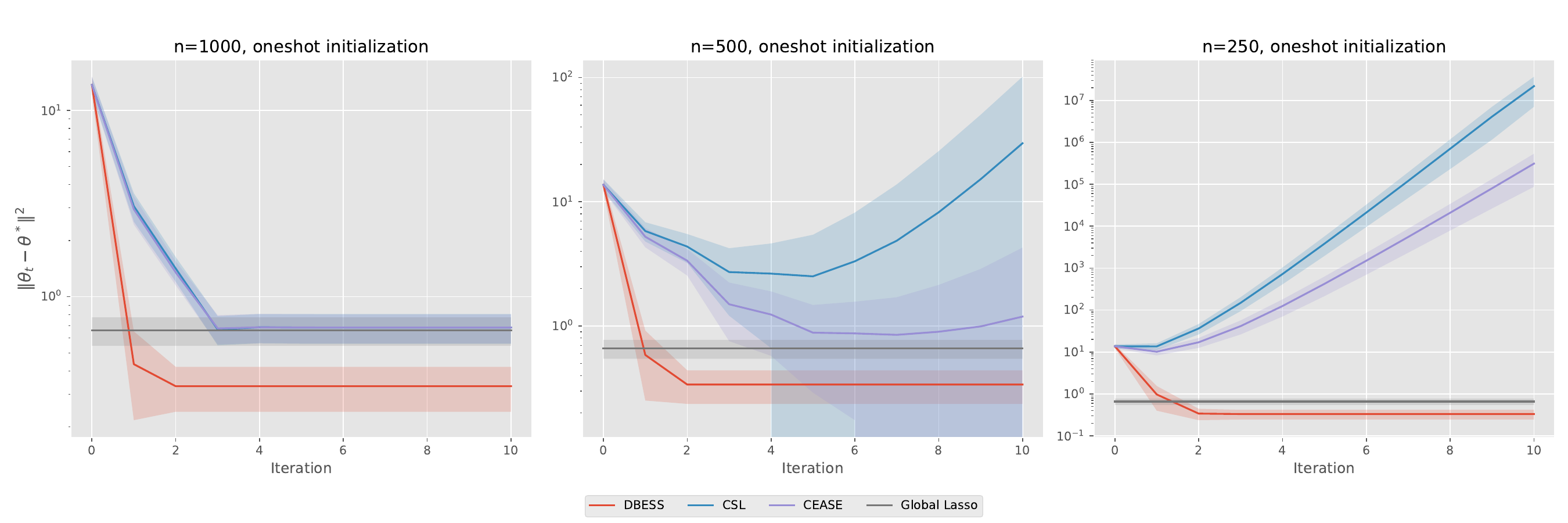}
    \end{subfigure}
    \\
    \begin{subfigure}[b]{\linewidth}
        \includegraphics[width=\linewidth]{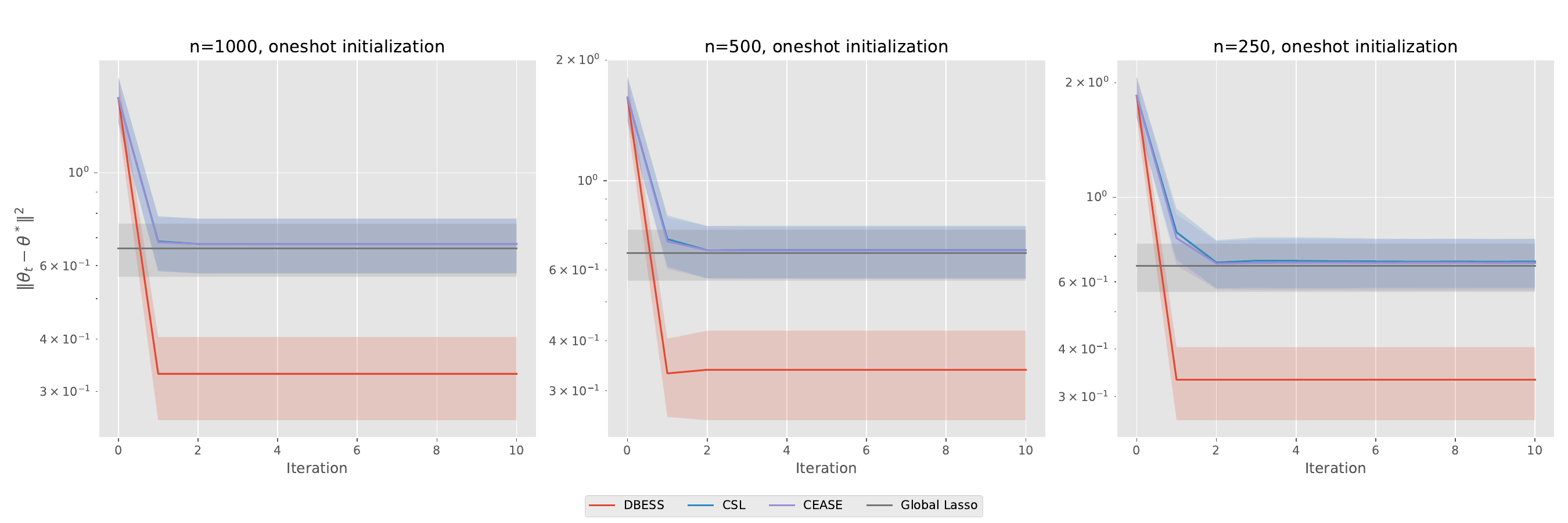}
    \end{subfigure}

    \caption{Comparative analysis of convergence for distributed regression algorithms. The x-axis represents the number of iterations, and the y-axis quantifies the estimation error $\|\boldsymbol{\theta}_t - \boldsymbol{\theta}^*\|_2$. The top and bottom panels use $\mathbf{0}$ and $\btheta_{\mathrm{oneshot}}$  for initialization.}
    \label{f1}
\end{figure}
According to Figure \ref{f1}, we find that under oneshot initialization, DBESS, CSL, and CEASE algorithms can all converge, but the relative error of the DBESS algorithm decreases faster. Another point is that the errors of CSL and CEASE at convergence are similar to the error of the global lasso, but DBESS obtains a smaller error and a more accurate solution. This phenomenon is intuitive. The estimates obtained based on the $l_1$ penalty are biased, and the accuracy of the solutions obtained by the CSL and CEASE methods will not be better than that of the global lasso. DBESS accurately recovers sparsity and the active set and estimates the active solution, so the error is naturally smaller.

Moreover, with zero initialization, the convergence behaviors of CSL, CEASE, and DBESS algorithms exhibit notable differences. All three methods achieve convergence for a relatively sizeable per-machine sample size (\(n=1000\)). However, with reduced sample sizes, CSL (\(n=500\)) and CEASE (\(n=250\)) may fail to converge, whereas DBESS continues to do so. In summary, DBESS converges more quickly and delivers more accurate solutions, demonstrating its effectiveness with smaller sample sizes per machine. The effect of the number of machines and local sample size will be further investigated in Section 4.4.

\subsection{Recovery of sparse subsets}

In this study, we analyze the efficacy of various algorithms for high-dimensional sparse linear regression models across different distributed settings. Our assessment centers on accurately identifying the true active set, leveraging key metrics such as TPR, TNR, MCC, and ReEE, as elaborated in Section 4.1.

Our analysis explores a scenario with parameters $(N, m, p, s^*) = (10000, 10, 100, 10)$. We investigate three covariance matrix structures mentioned in Section 4.1: uncorrelated, low correlation, and high correlation. To enhance algorithm performance, we employ oneshot initialization. The estimates produced by the CSL, CEASE methods may not inherently exhibit sparsity. To address this, we employ the truncation technique mentioned in \cite{Battey2018}, setting coefficients with absolute values smaller than \(\nu = 3\sqrt{\frac{\log p}{N}}\) to zero.
Table \ref{tab:1} displays the comparative performance of these algorithms.

\begin{table}[H]
\centering
\caption{\textbf{TPR, TNR, MCC, Relative errors for different number of machines}}
\label{tab:1}
\begin{tabular}{lcccc}
\hline
\textbf{Method} & \textbf{TPR} & \textbf{TNR} & \textbf{MCC} & \textbf{ReEE} \\
\hline
& & $m=20$ & &\\
CEASE & \textbf{1.000(0.000)} & 0.826 (0.066) & 0.581 (0.089) & 0.093 (0.020) \\
CSL & \textbf{1.000 (0.000)} & 0.827 (0.061) & 0.580 (0.083) & 0.093 (0.020) \\
DBESS & \textbf{1.000 (0.000)} & \textbf{1.000 (0.001)} & \textbf{0.999 (0.005)} & \textbf{0.038 (0.015)} \\
Global Lasso & \textbf{1.000 (0.000)} & 0.982 (0.015) & 0.925 (0.058) & 0.093 (0.020)  \\
\hline
& & $m=50$ & &\\
CEASE & \textbf{1.000(0.000)} & 0.820 (0.071) & 0.574 (0.090) & 0.096 (0.022) \\
CSL & \textbf{1.000 (0.000)} & 0.810 (0.103) & 0.570 (0.115) & 0.127 (0.062) \\
DBESS & \textbf{1.000 (0.000)} & \textbf{1.000 (0.002)} & \textbf{0.998 (0.010)} & \textbf{0.039 (0.015)} \\
Global Lasso & \textbf{1.000 (0.000)} & 0.982 (0.015) & 0.925 (0.058) & 0.093 (0.020)  \\
\hline
& & $m=80$ & &\\
CEASE & 0.998 (0.020) & 0.638 (0.166) & 0.409 (0.132) & 0.201 (0.189) \\
CSL & 0.980 (0.045) & 0.090 (0.152) & 0.053 (0.109) & 263.716 (604.583) \\
DBESS & \textbf{1.000 (0.000)} & \textbf{0.998 (0.004)} & \textbf{0.992 (0.019)} & \textbf{0.041 (0.016)} \\
Global Lasso & \textbf{1.000 (0.000)} & 0.982 (0.015) & 0.925 (0.058) & 0.093 (0.020) \\
\hline
\end{tabular}

\end{table}

All the evaluated methods demonstrated high TPR, affirming their capability to identify the true variables. However, the DBESS method exhibited a higher TNR, effectively reducing the likelihood of including irrelevant variables. This characteristic, coupled with its substantial TPR, suggests that DBESS can efficiently select all the true variables without incorporating excessive predictors. Despite the application of shrinkage, the CSL, CEASE, and Global Lasso methods still did not achieve high TNRs, indicating a tendency to select more variables and, consequently, perform less impressively in terms of the MCC. DBESS showed the smallest ReEE, implying that it provides the most precise estimates. Moreover, we observed that the performance of the algorithms is significantly influenced by the number of machines utilized in the computations. When the number of machines is relatively small (e.g., \(m=20\)), methods such as CSL and CEASE demonstrate lower relative errors. However, as the number of machines increases (e.g., \(m=80\)), the error associated with the CEASE method increases noticeably, and the CSL method exhibits convergence issues. In contrast, the DBESS method consistently identifies key variables and provides accurate estimates, maintaining its effectiveness even with many machines and smaller local sample sizes. In summary, DBESS emerges as a robust and effective approach for variable selection and estimation in distributed high-dimensional environments, offering precise recovery of the active set, accurate estimations, and stability across different settings.

\subsection{Effect of Machine Number and Local Sample Size}
This section explores how variations in the number of machines and the local sample sizes affect the algorithm's estimation accuracy. Our investigation is bifurcated into two distinct experimental setups: in the first, we fix the local sample size and vary the number of machines; in the second, we maintain a consistent number of machines while adjusting local sample sizes. Our goal is to evaluate and contrast the squared estimation errors(SEE), expressed as $\|\boldsymbol{\theta}-\boldsymbol{\theta}^*\|^2$, across these configurations to ascertain their impact on algorithmic efficacy.

For this analysis, we specify the feature dimension and sparsity level as $(p,s^*)=(80,8)$ and adopt an uncorrelated covariance matrix $\bSigma$. Figure \ref{f3} illustrates the outcomes of these investigations. The left graph presents the effects of altering the number of machines $m$ within the set $\{ 1, 5, 10, 20, 30, 40, 50\}$, holding the local sample size at $n=200$. The right graph, conversely, showcases the influence of varying local sample sizes $n$ in the range of $\{150, 200, 250, 300, 350, 400, 450\}$ while keeping the number of machines constant at $m=30$. Both scenarios employ a one-shot initialization strategy for setting initial parameter estimates.

From the left graph of Figure \ref{f3}, it is observed that with a fixed local sample size ($n$), increasing the number of machines ($m$) generally leads to a decrease in the squared estimation errors for various algorithms. However, the CSL algorithm begins to diverge from the global Lasso, exhibiting more significant errors, while the CEASE  algorithm performs intermediately between the two. This observation aligns with theoretical analysis. For CSL to achieve precision comparable to the global model, it requires $n > m(s^*)^2 \log p$. As $m$ increases, fulfilling this criterion becomes challenging, leading to a growing discrepancy. CEASE, through applying the PPA, somewhat relaxes this requirement, hence showing a slightly improved performance over CSL. On the other hand, DBESS  achieves a smaller SEE than global Lasso by accurately recovering the active set, devoid of the inherent bias associated with Lasso. In the right graph, where the number of machines is fixed, a similar trend of differences in SEE among CSL, CEASE, and Global Lasso is initially evident when $n$ is small. Still, the performance of these three methods converges as $n$ increases. DBESS continues to exhibit the smallest SEE.

DBESS demonstrates a relatively lower dependency on both the local sample size and the number of machines, achieving more precise estimates under similar settings.

\begin{figure}[H]
    \centering
    \includegraphics[width=\textwidth]{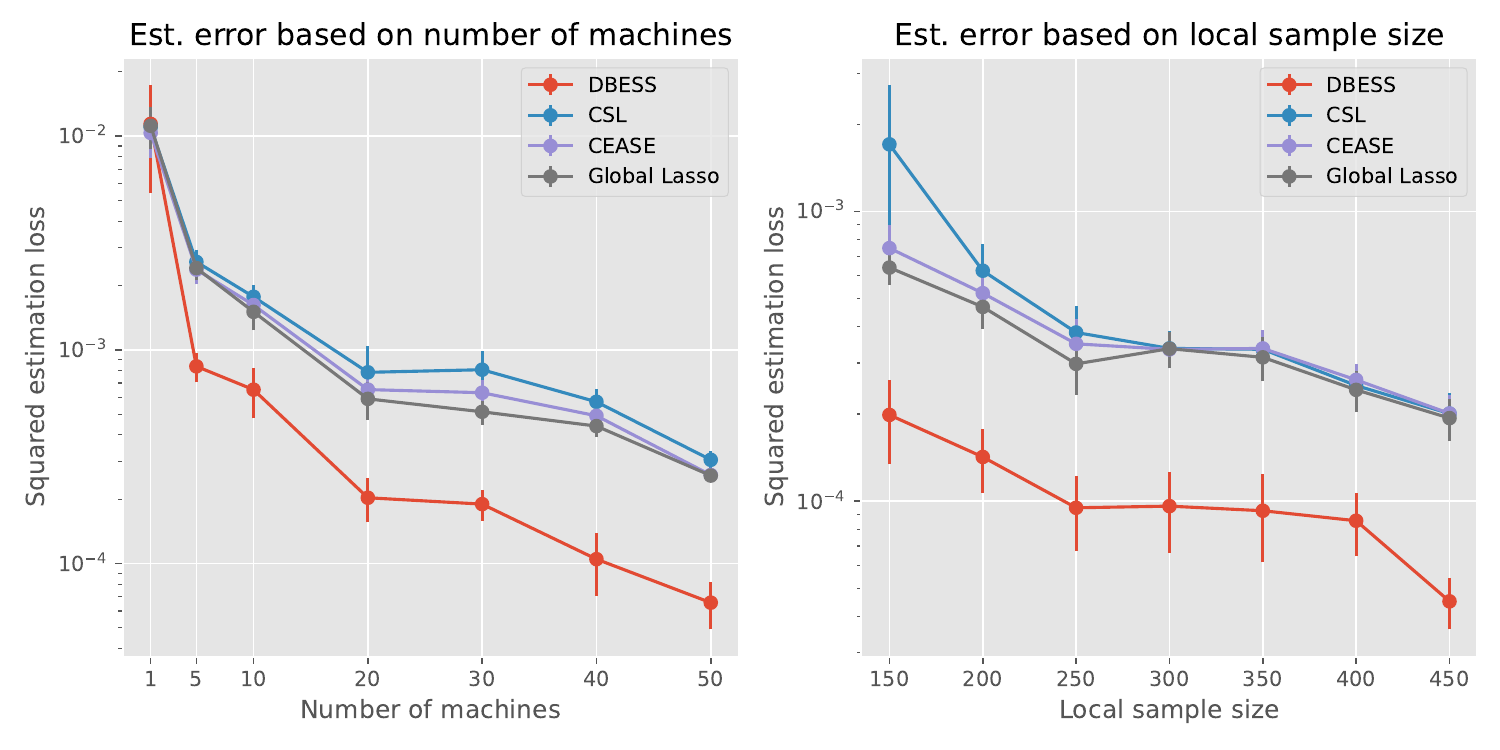}
    \caption{Mean squared estimation error as a function of the number of machines ($m$) and local sample size ($n$). The left graph illustrates the impact of varying $m$ (within $\{1, 5, 10, 20, 30, 40, 50\}$) while holding $n=200$ constant. The right graph explores how changes in $n$ (across $\{150, 200, 250, 300, 350, 400, 450\}$), with a fixed $m=30$, affect the error.}
    \label{f3}
\end{figure}

\subsection{Real Data Analysis}
\label{fifth_real}
For a real data example, we use the Communities and Crime dataset \citep{redmond2002data}, found in the UCI Machine Learning Repository. This dataset integrates socioeconomic data, law enforcement information, and crime statistics from multiple sources in the United States for the years 1990 and 1995. After removing missing values, we have 1,993 observations and 101 variables. We use the total number of violent crimes per 100,000 people (ViolentCrimesPerPop) as the response variable and select the remaining 99 continuous variables as predictors. We randomly split the dataset into training and test sets with an $80/20$ ratio. In the distributed algorithm, we set the number of machines to $m=4$, initialized with the one-shot average. We then compare the algorithms based on their mean squared error (MSE) on the test set. Figure \ref{f3} shows the average performance of DBESS, CSL, CEASE, and Global Lasso. We found that the DBESS algorithm converged in four iterations and achieved a lower MSE than the Global Lasso. In contrast, the CEASE algorithm required more iterations to converge, with an MSE close to that of the Global Lasso. The CSL method was also unstable; although we observed a decrease in MSE during the initial iterations, it did not converge in subsequent iterations.
\begin{figure}[ht]
    \centering
    \includegraphics[width=\linewidth]{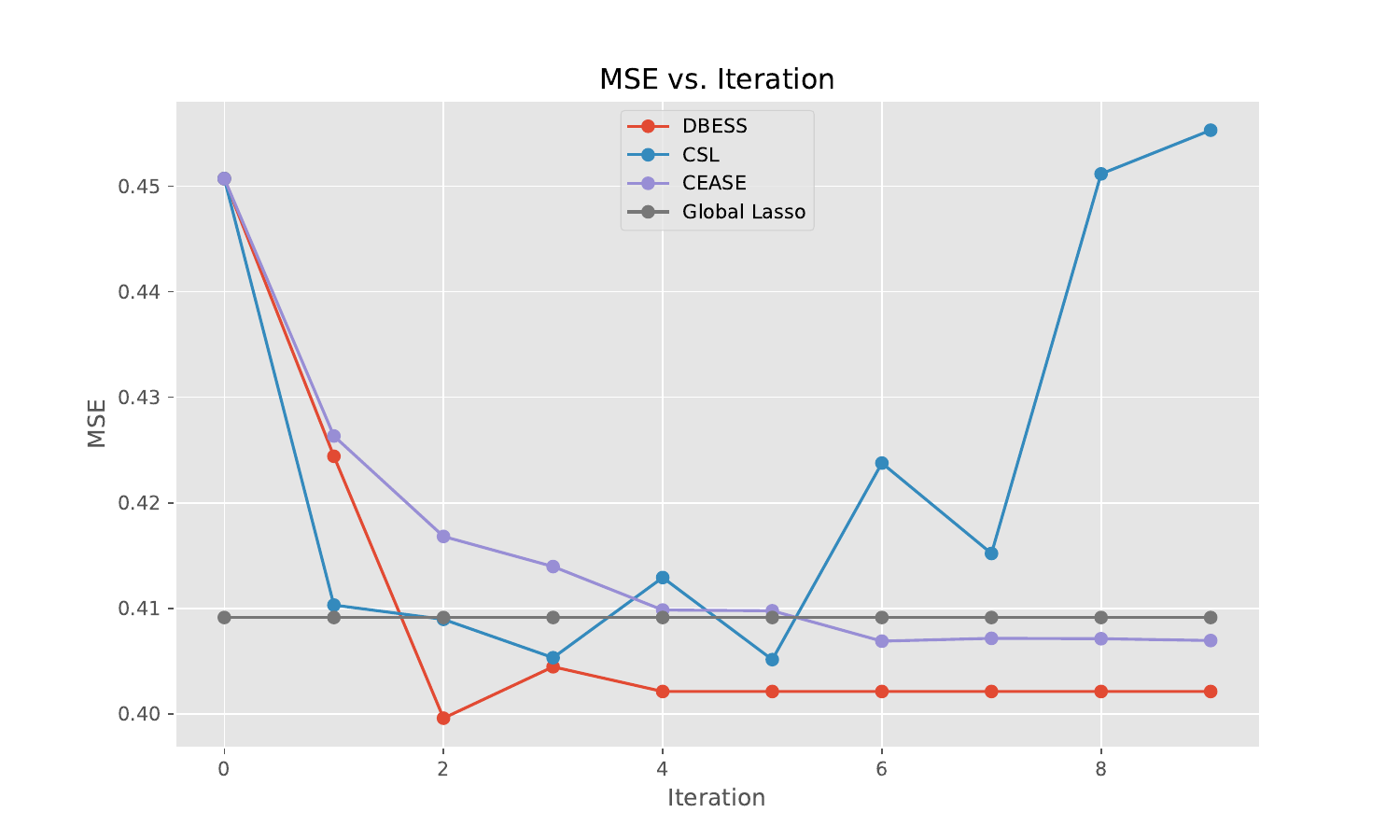}
    \caption{Comparative analysis of convergence for distributed regression algorithms on Communities and Crime dataset. The $x$-axis represents the number of iterations, and the $y$-axis shows the test set's mean squared error (MSE). The algorithms are initialized with one-shot average  $\btheta_{\mathrm{oneshot}}$.}
    \label{f3}
\end{figure}

The DBESS algorithm, unlike the CEASE and CSL algorithms, which failed to generate sparse estimates, demonstrated superior performance on the test set and, notably, selected only eight variables. These variables include total population (population), urban population (numbUrban), percentage of children in two-parent families (PctKids2Par), percentage of illegitimate children (PctIlleg), percentage of people living in crowded housing (PctPersDenseHous), and the number of vacant houses (HousVacant). These factors are crucial socioeconomic indicators reflecting various influences on community crime rates. A negative correlation between total population and violent crime rates suggests that larger communities may have advantages in resource distribution and social control mechanisms. Conversely, an increase in urban population correlates positively with higher crime rates, indicating that urban anonymity and loose community ties may facilitate criminal activities. Family structure significantly impacts crime rates: communities with a higher proportion of children from two-parent families generally exhibit lower crime rates, whereas communities with a higher percentage of illegitimate children experience higher crime rates. Additionally, housing conditions are also linked to crime rates. Both overcrowded living conditions and an increase in the number of vacant houses correlate with higher crime rates, reflecting the influence of economic pressures and community neglect on criminal activity.

\section{Conclusion}
\label{sec-con}
In this paper, we propose a two-stage distributed learning algorithm with high communication efficiency, aiming to handle massive data distributed across multiple machines for best subset selection of high-dimensional sparse linear models. We further design a data-based information criterion for adaptive sparsity selection, which ensures the consistency of variable selection, effectively avoids the problems of over-selection and omission of variables, and accurately recovers the real sparse model. Numerical experiments verify the method's convergence speed and estimation accuracy advantages.

Of course, there are still numerous directions worth exploring in the follow-up work of this study. Firstly, while we have implemented a basic one-shot average strategy for parameter estimation within the identified active set, future efforts will focus on developing more effective estimation techniques. Furthermore, we plan to extend the present algorithm to decentralized distributed settings and apply it to generalized linear models. Additionally, interval estimation and hypothesis testing in distributed environments are crucial research directions. We are confident that the framework proposed in this study, which prioritizes identifying the active set of high-dimensional sparse parameters before estimation and inference, will find broad applicability in distributed learning.

\bibliographystyle{agsm}

\bibliography{reference}

@article{chen2023communication,
  title={Communication-efficient estimation for distributed subset selection},
  author={Chen, Yan and Dong, Ruipeng and Wen, Canhong},
  journal={Statistics and Computing},
  volume={33},
  number={6},
  pages={141},
  year={2023},
  publisher={Springer}
}

@article{redmond2002data,
  title={A data-driven software tool for enabling cooperative information sharing among police departments},
  author={Redmond, Michael and Baveja, Alok},
  journal={European Journal of Operational Research},
  volume={141},
  number={3},
  pages={660--678},
  year={2002},
  publisher={Elsevier}
}

@article{wang2007tuning,
  title={Tuning parameter selectors for the smoothly clipped absolute deviation method},
  author={Wang, Hansheng and Li, Runze and Tsai, Chih-Ling},
  journal={Biometrika},
  volume={94},
  number={3},
  pages={553--568},
  year={2007},
  publisher={Oxford University Press}
}

@article{huang2019distributed,
  title={A distributed one-step estimator},
  author={Huang, Cheng and Huo, Xiaoming},
  journal={Mathematical Programming},
  volume={174},
  pages={41--76},
  year={2019},
  publisher={Springer}
}

@article{Raskutti2011minimax,
  title={Minimax rates of estimation for high-dimensional linear regression over $l_q$-balls},
  author={Raskutti, Garvesh and Wainwright, Martin J and Yu, Bin},
  journal={IEEE transactions on information theory},
  volume={57},
  number={10},
  pages={6976--6994},
  year={2011},
  publisher={IEEE}
}

@article{Battey2018,
  title = {Distributed testing and estimation under sparse high dimensional models},
  author = {Battey, H. and Fan, J. and Liu, H. and Lu, J. and Zhu, Z. },
  year = {2018},
  journal = {The Annals of Statistics},
  volume = {46},
  number = {},
  pages = {1352--1382},
  publisher = {{Institute of Mathematical Statistics}},
  urldate = {2023-05-29}
}

@article{Chen2022FirstOrder,
  title = {First-{{Order Newton-Type Estimator}} for {{Distributed Estimation}} and {{Inference}}},
  author = {Chen, Xi and Liu, Weidong and Zhang, Yichen},
  year = {2022},
  journal = {Journal of the American Statistical Association},
  volume = {117},
  number = {540},
  pages = {1858--1874},
  publisher = {{Taylor \& Francis}},
  urldate = {2023-02-23}
}

@article{Fan2001,
  title = {Variable selection via nonconcave penalized likelihood and its oracle properties},
  author = {Fan, J. and Li, R.},
  year = {2001},
  journal = {Journal of the American statistical Association},
  volume = {96},
  number = {456},
  pages = {1348--1360},
  urldate = {2023-07-03},
  langid = {english}
}

@article{Fan2013Tuning,
  title = {Tuning {{Parameter Selection}} in {{High Dimensional Penalized Likelihood}}},
  author = {Fan, Yingying and Tang, Cheng Yong},
  year = {2013},
  journal = {Journal of the Royal Statistical Society Series B: Statistical Methodology},
  volume = {75},
  number = {3},
  pages = {531--552},
  urldate = {2023-04-12},
  langid = {english}
}

@article{Fan2021CommunicationEfficient,
  title = {Communication-{{Efficient Accurate Statistical Estimation}}},
  
author = {Fan, Jianqing and Guo, Yongyi and Wang, Kaizheng},
  year = {2023},
  journal = {Journal of the American Statistical Association},
  volume = {118},
  number = {542},
  pages = {1000--1010},
  publisher = {{Taylor \& Francis}},
  urldate = {2023-02-23}
}

@misc{Gu2022Weighted,
  title = {Weighted {{Distributed Estimation}} under {{Heterogeneity}}},
  author = {Gu, Jia and Chen, Songxi},
  year = {2022},
  number = {arXiv:2209.06482},
  eprint = {2209.06482},
  primaryclass = {math, stat},
  publisher = {{arXiv}},
  urldate = {2023-05-31},
  archiveprefix = {arxiv}
}

@article{Huang2018Constructive,
  title = {A {{Constructive Approach}} to \${{L}}\_0\$ {{Penalized Regression}}},
  author = {Huang, Jian and Jiao, Yuling and Liu, Yanyan and Lu, Xiliang},
  year = {2018},
  journal = {Journal of Machine Learning Research},
  volume = {19},
  number = {10},
  pages = {1--37},
  urldate = {2023-04-17}
}

@article{Jordan2019CommunicationEfficient,
  title = {Communication-{{Efficient Distributed Statistical Inference}}},
  author = {Jordan, Michael I. and Lee, Jason D. and Yang, Yun},
  year = {2019},
  journal = {Journal of the American Statistical Association},
  volume = {114},
  number = {526},
  pages = {668--681},
  publisher = {{Taylor \& Francis}},
  urldate = {2023-01-18}
}

@article{Lee2017Communicationefficient,
  title = {Communication-Efficient {{Sparse Regression}}},
  author = {Lee, Jason D. and Liu, Qiang and Sun, Yuekai and Taylor, Jonathan E.},
  year = {2017},
  journal = {Journal of Machine Learning Research},
  volume = {18},
  number = {5},
  pages = {1--30},
  urldate = {2023-04-16}
}

@inproceedings{Liu2014Distributed,
  title = {Distributed {{Estimation}}, {{Information Loss}} and {{Exponential Families}}},
  booktitle = {Advances in {{Neural Information Processing Systems}}},
  author = {Liu, Qiang and Ihler, Alexander T},
  year = {2014},
  volume = {27},
  publisher = {{Curran Associates, Inc.}},
  urldate = {2023-05-29}
}

@inproceedings{Shamir2014CommunicationEfficient,
  title = {Communication-{{Efficient Distributed Optimization}} Using an {{Approximate Newton-type Method}}},
  booktitle = {Proceedings of the 31st {{International Conference}} on {{Machine Learning}}},
  author = {Shamir, Ohad and Srebro, Nati and Zhang, Tong},
  year = {2014},
  pages = {1000--1008},
  publisher = {{PMLR}},
  urldate = {2023-03-17},
  langid = {english}
}

@article{Tibshirani1996,
  title = {Regressio shrinkage and selection via the lasso},
  author = {Tibshirani,R.},
  year = {1996},
  journal = {Journal of the Royal Statistical Society, Series B},
  volume = {0},
  number = {0},
  pages = {267--288},
  publisher = {{Taylor \& Francis}},
  urldate = {2023-07-03}
}

@inproceedings{Wang2017Efficient,
  title = {Efficient {{Distributed Learning}} with {{Sparsity}}},
  booktitle = {Proceedings of the 34th {{International Conference}} on {{Machine Learning}}},
  author = {Wang, Jialei and Kolar, Mladen and Srebro, Nathan and Zhang, Tong},
  year = {2017},
  pages = {3636--3645},
  publisher = {{PMLR}},
  urldate = {2023-02-23},
  langid = {english}
}

@inproceedings{Zhang2013ComunicationEfficient,
  title = {Comunication-{{Efficient Algorithms}} for {{Statistical Optimization}}},
  booktitle = {Advances in {{Neural Information Processing Systems}}},
  author = {Zhang, Yuchen and Duchi, John C. and Wainwright, Martin},
  year = {2012},
  volume = {25},
  publisher = {{Curran Associates, Inc.}},
  urldate = {2023-05-29}
}

@article{Zhu2020Polynomial,
  title = {A {{Polynomial Algorithm}} for {{Best-Subset Selection Problem}}},
  author = {Zhu, Junxian and Wen, Canhong and Zhu, Jin and Zhang, Heping and Wang, Xueqin},
  year = {2020},
  journal = {Proceedings of the National Academy of Sciences},
  volume = {117},
  number = {52},
  pages = {33117--33123},
  publisher = {{Proceedings of the National Academy of Sciences}},
  urldate = {2023-03-09}
}

@article{Zhu2021LeastSquare,
  title = {Least-{{Squares Approximation}} for a {{Distributed System}}},
  author = {Zhu, Xuening and Li, Feng and Wang, Hansheng},
  year = {2021},
  journal = {Journal of Computational and Graphical Statistics},
  volume = {30},
  number = {4},
  pages = {1004--1018},
  publisher = {{Taylor \& Francis}},
  urldate = {2023-02-23}
}

@article{Zhu2023Fast,
  title = {A {{Fast Algorithm}} to {{Best-Subset Selection}} in {{Generalized Linear Models}}},
  author = {Zhu, Junxian and Zhu, Jin and Tang, Borui and Chen, Xuanyu and Wang, Xueqin},
  year = {2023},
  langid = {english}
}

\end{document}


\def\spacingset#1{\renewcommand{\baselinestretch}%
{#1}\small\normalsize} \spacingset{1}


\if1\blind
{
  \title{\bf Supplement to ``Communication-Efficient Distributed Best Subset Selection with Statistical Guarantees"}
  \author{Jingguo Lan\\
    University of Science and Technology of China\\
    and \\
    Hongmei Lin \thanks{Lin’s research is partially supported by the National Natural Science Foundation of China (12171310) and the Shanghai "Project Dawn 2022" (22SG52).}\hspace{.2cm}\\
    School of Statistics and Information, \\Shanghai University of International Business and Economics\\
    and \\
    Xueqin Wang \thanks{Wang’s research is partially supported by the National Natural Science Foundation of China (12231017, 72171216, 71921001, and 71991474) and the National Key R\&D Program of China (2022YFA1003800).}\hspace{.2cm}\\
    International Institute of Finance, School of Management, \\ University of Science and Technology of China}
    
  \maketitle
 \fi

\if0\blind
{
  \bigskip
  \bigskip
  \bigskip
  \begin{center}
    {\LARGE\bf Supplement to ``Minimax and Communication-Efficient Distributed Best Subset Selection with Oracle Property"}
\end{center}
  \medskip
} \fi

\bigskip
This supplementary material provides detailed proofs of lemmas and theorems mentioned in the article. Section \ref{secA} presents the proofs of several lemmas referred to in Section 2.2 of the manuscript. Section \ref{secB} elaborates on the proofs of several theorems stated in Section 3.2 of the manuscript. The main framework of the proof references \cite{Zhu2020Polynomial, Zhu2023Fast}.
\vfill

\section{Proofs of Lemmas}
\label{secA}

\subsection{Proof of Lemma 2.1}
Consider the optimization problem
\begin{equation}
\min_{\boldsymbol{\theta}} f(\boldsymbol{\theta})=\frac{1}{2} \boldsymbol{\theta}^T \bG \boldsymbol{\theta} + \boldsymbol{b}^T \boldsymbol{\theta}  
\quad \text{s.t. } \| \boldsymbol{\theta}\|_0=s,
\end{equation}
where $\|\boldsymbol{\theta}\|_0$ denotes the number of non-zero elements in the vector $\boldsymbol{\theta}$.\\
Denote augmented Lagrangian function as: 
$$
L_\rho (\boldsymbol{\theta}, \boldsymbol{v}, \boldsymbol{d})= \frac{1}{2} \boldsymbol{\theta}^T \bG \boldsymbol{\theta} + \boldsymbol{b}^T \boldsymbol{\theta} + \boldsymbol{d}^T(\boldsymbol{\theta}-\boldsymbol{v})+ \frac{\rho}{2}\|\boldsymbol{\theta} - \boldsymbol{v} \|^2.
$$
Consider Block coordinate-wise minimizer for $\boldsymbol{\theta}, \boldsymbol{v}, \boldsymbol{d}$  respectively.
\begin{itemize}
    \item   Given $(\boldsymbol{v},\boldsymbol{d})$, minimize over $\boldsymbol{\theta}$:\\
   By taking the partial derivative with respect to $\boldsymbol{\theta}$ and setting it to zero, we get:
   $$
   \frac{\partial L_{\rho}}{\partial \boldsymbol{\theta}}=\bG \boldsymbol{\theta}+\boldsymbol{b}+\boldsymbol{d}+\rho(\boldsymbol{\theta}-\boldsymbol{v})=0.
   $$

\item Given $(\boldsymbol{\theta},\boldsymbol{d})$, minimize over $\boldsymbol{v}$:\\
   Considering the constraint $\|\boldsymbol{v}\|_0=s$, we frame the problem as:
   $$
   \boldsymbol{v} = \argmin_{\|\boldsymbol{v}\|_0 = s}  \frac{\rho}{2} \left\| \left( \boldsymbol{\theta} + \frac{1}{\rho} \boldsymbol{d} \right) - \boldsymbol{v} \right\|_2^2.
   $$
   This means $\boldsymbol{v}$ should match the $s$ largest absolute components of $\boldsymbol{\theta} + \frac{1}{\rho}\boldsymbol{d}$ in the active set $\mathcal{A}$. Therefore, the active set $\mathcal{A}$ is defined as:
   \begin{equation}
   \mathcal{A}=\left\{j: \sum_{i=1}^p \mathbb{I}\left(\left|\boldsymbol{\theta}_j+\frac{1}{\rho} \boldsymbol{d}_j\right| \leqslant\left|\boldsymbol{\theta}_i+\frac{1}{\rho} \boldsymbol{d}_i\right|\right) \leqslant s\right\}.
   \end{equation}

\item  Given $(\boldsymbol{v}, \boldsymbol{\theta})$, minimize over $\boldsymbol{d}$:\\
   By taking the partial derivative with respect to $\boldsymbol{d}$ and setting it to zero, we obtain:
   $$
   \frac{\partial L_\rho}{\partial \boldsymbol{d}}=\boldsymbol{\theta}-\boldsymbol{v}=0.
   $$
\end{itemize}

Based on the analysis above, the optimal solution must satisfy the following conditions:
\begin{equation}
\begin{aligned}
    &\mathcal{A}^{\diamond} = \left\{j: \sum_{i=1}^p \mathbb{I}\left(\left|\boldsymbol{\theta}_j^{\diamond}+\frac{1}{\rho} \boldsymbol{d}_j^{\diamond}\right| \leqslant\left|\boldsymbol{\theta}_i^{\diamond}+\frac{1}{\rho} \boldsymbol{d}_i^{\diamond}\right|\right) \leqslant s\right\}, \qquad
    \mathcal{I}^{\diamond} = (\mathcal{A}^{\diamond})^c, \\
    &\boldsymbol{\theta}^{\diamond}_{\mathcal{A}^{\diamond}} = -\left(\boldsymbol{G}_{\mathcal{A}^{\diamond}\times \mathcal{A}^{\diamond}}\right)^{-1}\boldsymbol{b}_{\mathcal{A}^{\diamond}}, \qquad
    \boldsymbol{\theta}^{\diamond}_{\mathcal{I}^{\diamond}} =\boldsymbol{0}, \\
    &\boldsymbol{d}^{\diamond}_{\mathcal{A}^{\diamond}} = \boldsymbol{0}, \qquad
    \boldsymbol{d}^{\diamond}_{\mathcal{I}^{\diamond}} = -(\boldsymbol{G}_{\mathcal{I}^{\diamond}\times \mathcal{I}^{\diamond}}) \boldsymbol{\theta}^{\diamond}_{\mathcal{I}^{\diamond}}-\boldsymbol{b}_{\mathcal{I}^{\diamond}}, \\
    &\boldsymbol{v}^{\diamond} = \boldsymbol{\theta}^{\diamond}.
\end{aligned}
\end{equation}

\subsection{Proof of Lemma 2.2}
Consider the loss function $l_n(\boldsymbol{\theta}) $, which is quadratic in $\boldsymbol{\theta} $. We define $\hat{\boldsymbol{\theta}} $ as the minimizer of $l_n $ under the constraint that $\boldsymbol{\theta}_{\mathcal{A}^c} = \boldsymbol{0} $,
i.e., $\hat \btheta = \argmin_{\btheta_{\cA^c}=\boldsymbol{0}} l_n(\btheta)$.
Given this setup, we have the critical point condition:
$$
    \left(\nabla_{\cA} l_n(\hat\btheta)\right) ^{\top}\hat \btheta_{\cA} =\left(\bb + \bG \hat \btheta\right)_{\cA}^{\top} \hat\btheta_{\cA}=\boldsymbol{0}. \label{eqp2.2.1}
$$
Since $l_n(\btheta)$ is a quadratic function, it can be expanded at $\hat{\boldsymbol{\theta}}$ as:
$$
\begin{aligned}
l_n(\btheta) =  l(\hat{\btheta})+\left(\bb+ \bG \hat \btheta\right)^{\top}\left({\btheta}-\hat{\btheta}\right)
 +\frac{1}{2}\left({\btheta}-\hat{\btheta}\right)^{\top} \bG \left({\btheta}-\hat{\btheta}\right).
\end{aligned}
$$
For any $ j \in \mathcal{A} $, considering the alteration of $ \hat{\boldsymbol{\theta}} $ by excluding $ j $ (denoted as $ \hat{\boldsymbol{\theta}}|_{\mathcal{A} \backslash\{j\}} $), we have:
$$
\begin{aligned}
l_n\left(\hat{\btheta}|_{\mathcal{A} \backslash\{j\}}\right)-l_n(\hat{\btheta})
&= \left(\bb+ \bG \hat \btheta\right)^{\top}\left(\hat{\btheta}|_{\mathcal{A} \backslash\{j\}}-\hat{\btheta}\right)
 +\frac{1}{2}\left(\hat{\btheta}|_{\mathcal{A} \backslash\{j\}}-\hat{\btheta}\right)^{\top} \bG \left(\hat{\btheta}|_{\mathcal{A} \backslash\{j\}}-\hat{\btheta}\right) \\
& =- \left(\nabla l_n(\hat{\btheta})\right)_j \hat \theta_j
 +\frac{1}{2} G_{jj} (\hat{\theta}_j^2)\\
 &= \frac{1}{2} G_{jj} (\hat{\theta}_j^2).
\end{aligned}
$$
The last equation is based on E.q.(\ref{eqp2.2.1}). $\left(\nabla l_n(\hat{\btheta})\right)_j \hat \theta_j =0 $, for $j \in \cA$. \\
For $ j \notin \mathcal{A} $, we introduce a perturbation $ t \boldsymbol{1}_j $ to $ \hat{\boldsymbol{\theta}} $ and analyze the effect on the loss function:
$$
\begin{aligned}
l_n\left(\hat{\btheta}+ t \boldsymbol{1}_j\right)-l_n(\hat{\btheta})
&= \left(\bb+ \bG \hat \btheta\right)^{\top}\left(\hat{\btheta}+ t \boldsymbol{1}_j-\hat{\btheta}\right)
 +\frac{1}{2}\left(\hat{\btheta}+ t \boldsymbol{1}_j-\hat{\btheta}\right)^{\top} \bG \left(\hat{\btheta}+ t \boldsymbol{1}_j-\hat{\btheta}\right) \\
& =- t \hat d_j + \frac{1}{2} t^2 G_{jj},
\end{aligned}
$$
where $\hat \bd= \nabla l_n(\hat \btheta)=\bb + \bG \hat \btheta.$
Note that the right side of the above formula is a quadratic function in $t$. With simple operations, the maximum value can be obtained as $\frac{1}{2}[G_{jj}]^{-1} \hat d_j^2$.

\subsection{Proof of Lemma 2.3}

Given that $C<|\mathcal{A}^k|$, we consider the following inequality:
$$
\min _{j \in \mathcal{S}_{C, 2}^k} \frac{1}{\rho^2}\left|d_{j}^k\right|^2 = \min _{j \in \mathcal{S}_{C, 2}^k}\left|\beta_{j}^k + \frac{1}{\rho} d_{j}^k\right|^2 \geqslant \max _{i \in \mathcal{A}^k}\left|\theta_{i}^k + \frac{1}{\rho} d_{i}^k\right|^2 = \max _{i \in \mathcal{A}^k}\left|\theta_{i}^k\right|^2 \geqslant \max _{i \in \mathcal{S}_{C, 1}^k}\left|\theta_{i}^k\right|^2.
$$
Through algebraic manipulation, we derive the corresponding range for $\rho$ as:
\begin{equation} \label{eq3.2.1}
    \rho \leqslant \frac{\min _{j \in \mathcal{S}_{C, 2}^k}\left|d_{j}^k\right|}{\max _{i \in \mathcal{S}_{C, 1}^k}\left|\theta_{i}^k\right|}.
\end{equation}
Similar to (\ref{eq3.2.1}), given $C+1$,  the range of $\rho$ is
\begin{equation} \label{eq3.2.2}
    \rho \leqslant \frac{\min _{j \in \mathcal{S}_{C+1, 2}^k}\left|d_{j}^k\right|}{\max _{i \in \mathcal{S}_{C+1, 1}^k}\left|\theta_{i}^k\right|}.
\end{equation}
Notably, $\mathcal{S}_{C, 1}^k \subseteq \mathcal{S}_{C+1,1}^k$ and $\mathcal{S}_{C, 2}^k \subseteq \mathcal{S}_{C+1,2}^k$. If the restricted active and inactive sets happen to exchange \( C \) elements, the range of values for \(\rho\) should be the difference between (\ref{eq3.2.1}) and (\ref{eq3.2.2}). Consequently, for $C<|\mathcal{A}^k|$, the range for $\rho$ is specified as:
\begin{equation}
    \rho \in \left(\frac{\min _{j \in \mathcal{S}_{C+1,2}^k}\left|d_{j}^k\right|}{\max _{i \in \mathcal{S}_{C+1,1}^k}\left|\theta_{i}^k\right|}, \frac{\min _{j \in \mathcal{S}_{C, 2}^k}\left|d_{j}^k\right|}{\max _{i \in \mathcal{S}_{C, 1}^k}\left|\theta_{i}^k\right|}\right].
\end{equation}
For the case where $C=|\mathcal{A}^k|$, it follows that $\mathcal{S}_{C, 1}^k=\mathcal{A}^k$. Thus, we have:
$$
\min _{j \in \mathcal{S}_{C, 2}^k} \frac{1}{\rho^2}\left|d_{j}^k\right|^2 = \min _{j \in \mathcal{S}_{C, 2}^k}\left|\theta_{j}^k + \frac{1}{\rho} d_{j}^k\right|^2 \geqslant \max _{i \in \mathcal{A}^k}\left|\theta_{i}^k + \frac{1}{\rho} d_{i}^k\right|^2 = \max _{i \in \mathcal{A}^k}\left|\theta_{i}^k\right|^2.
$$
Accordingly, after simplifying the algebraic expressions, the range for $\rho$ is determined to be:
$$
\rho \in \left(0, \frac{\min _{j \in \mathcal{S}_{C, 2}^k}\left|d_{j}^k\right|}{\max _{i \in \mathcal{A}^k}\left|\theta_{i}^k\right|}\right].
$$

\section{Proofs of Theorems}
\label{secB}

\subsection{Proof of Theorem 3.1}

\begin{proof}
Denote by $(\hat{\btheta}, \hat{\mathcal{A}})$ the solution of Algorithm 1 with a given support size $s$. That is,
$$
\hat\btheta= \argmin_{\|\btheta\|_0=s} l(\btheta)= \frac{1}{2}\btheta^{\top}\bSigma_1 \btheta- \bw^{\top} \btheta + \btheta_0^{\top}(\bSigma-\bSigma_1)\btheta.
$$
By substituting $\btheta_0 =\btheta^*+\bdelta$ 
and $\bw=\frac{1}{N} \bX^{\top} \bY=\frac{1}{N} \bX^{\top}(\bX\btheta^*+\bepsilon)$ into $l(\btheta)$, we get
$$
 l(\btheta)= \frac{1}{2} \btheta ^{\top} \bSigma_1 \btheta - [\bSigma \btheta^* +\frac{1}{N}\bX^{\top} \bepsilon]^{\top} \btheta + \bdelta^{\top}(\bSigma-\bSigma_1)\btheta.
$$
The gradient is
$$
\nabla l(\btheta)=\bSigma_1 \btheta - \bSigma_1 \btheta^* - \frac{1}{N}\bX^{\top} \bepsilon + (\bSigma-\bSigma_1)\bdelta.
$$
In particular, 
\begin{equation} \label{eq3.1.1}
    \nabla l(\btheta^*)=-\frac{1}{N}\bX^{\top} \bepsilon + (\bSigma-\bSigma_1)\bdelta.
\end{equation}
We assume $\mathcal{I}_1 \neq \varnothing$ and show that it will lead to a contradiction. Under this assumption, the algorithm will not terminate at $\hat{\btheta}$.
Let $k=\left|\mathcal{I}_1\right|$. Denote the splicing set in the active and inactive sets, respectively, as
$$
\begin{aligned}
\mathcal{S}_1=\left\{j \in \hat{\mathcal{A}}: \sum_{i \in \hat{\mathcal{A}}} \mathrm{I}\left(\xi_j \geq \xi_i\right) \leq k_0\right\}, \quad  
\mathcal{S}_2=\left\{j \in \hat{\mathcal{I}}: \sum_{i \in \hat{\mathcal{I}}} \mathrm{I}\left(\zeta_j \leq \zeta_i\right) \leq k_0\right\}.
\end{aligned}
$$
 Let
  $$
  \begin{aligned}
  & \mathcal{A}_{11}=\mathcal{A}_1 \cap\left(\mathcal{S}_1\right)^c, \mathcal{A}_{12}=\mathcal{A}_1 \cap \mathcal{S}_1, \\
  & \mathcal{A}_{21}=\mathcal{A}_2 \cap\left(\mathcal{S}_1\right)^c, \mathcal{A}_{22}=\mathcal{A}_2 \cap \mathcal{S}_1,
  \end{aligned}
  $$
  and
  $$
  \begin{aligned}
  & \mathcal{I}_{11}=\mathcal{I}_1 \cap \mathcal{S}_2, \mathcal{I}_{12}=\mathcal{I}_1 \cap\left(\mathcal{S}_2\right)^c, \\
  & \mathcal{I}_{21}=\mathcal{I}_2 \cap \mathcal{S}_2, \mathcal{I}_{22}=\mathcal{I}_2 \cap\left(\mathcal{S}_2\right)^c.
  \end{aligned}
  $$
Now we apply the splicing procedure on $\hat{\mathcal{A}}$ and $\hat{\mathcal{I}}$ using exchange sets $\mathcal{S}_1$ and $\mathcal{S}_2$, leading to the new active set $\tilde{\mathcal{A}}=\left(\hat{\mathcal{A}} \backslash \mathcal{S}_1\right) \cup \mathcal{S}_2$ and inactive set $\tilde{\mathcal{I}}=(\tilde{\mathcal{A}})^c$. Let $\tilde{\boldsymbol{\theta}}=\arg \min_{\btheta_{\tilde I} =0}  l(\btheta).$

We assert that the following two inequalities, which will be proven subsequently, hold with high probability.
\begin{enumerate}
    \item \begin{equation}
        \| \boldsymbol{\theta}^*_{\mathcal{A}_{12}}  \|_2 \le (2+\Delta) \frac{\nu_s}{m_s} \| \boldsymbol{\theta}^*_{\mathcal{I}_{1}} \|_2;
\end{equation}
  \item  \begin{equation}
        m_s\| \boldsymbol{\theta}^*_{\mathcal{I}_{12}}  \|_2 \le 2(1+\Delta) (\nu_s+\frac{\nu_s}{m_s}) \| \boldsymbol{\theta}^*_{\mathcal{I}_{1}}  \|_2.  
\end{equation}
  
\end{enumerate}
Based on the definition of $ \hat \btheta $ and the optimality conditions, we have:
$$
\boldsymbol{0}=\hat \bd_{\hat \cA}= \nabla_{\hat \cA}  l(\hat{\boldsymbol{\theta}}) = \nabla_{\hat \cA}  l(\boldsymbol{\theta}^*)+ \bSigma_{1,\hat \cA \times \hat \cA} (\hat{\boldsymbol{\theta}}_{\hat \cA} - \boldsymbol{\theta}^*_{\hat \cA}) + \bSigma_{1,\hat \cA \times \hat \cI_1} (-\boldsymbol{\theta}^*_{\hat \cI_1}). 
$$
Then,
\begin{equation}
    \|\hat \btheta_{\hat \cA}-\btheta^*_{\hat \cA}\|_2 \le 
\|  \bSigma_{1,\hat \cA \times \hat \cA}^{-1} (\nabla_{\hat \cA} l(\btheta^*)+ \bSigma_{1,\hat \cA \times  I_1} \boldsymbol{\theta}^*_{\hat \cI_1} )\|_2 
\le \frac{1}{m_s} \| \nabla_{\hat \cA} l(\btheta^*)\|_2+ \frac{\nu_s}{m_s}\| \boldsymbol{\theta}^*_{ \cI_1} \|_2. \label{eq3.1.0}
\end{equation}
The objective function $l(\boldsymbol{\theta})$ can be expanded around the point $\tilde{\boldsymbol{\theta}}$ in the following manner:
\begin{align}
l(\tilde{\boldsymbol{\theta}}) = &l\left(\boldsymbol{\theta}^*\right)+\left(\nabla_{\tilde{\mathcal{A}}} l(\boldsymbol{\theta^*})\right)^{\top}\left(\tilde{\boldsymbol{\theta}}_{\tilde{\mathcal{A}}}-\boldsymbol{\theta}_{\tilde{\mathcal{A}}}^*\right)+\left(\nabla_{\tilde{\mathcal{I}}_1} l(\boldsymbol{\theta^*})\right)^{\top}\left(-\boldsymbol{\theta}_{\tilde{\mathcal{I}}_1}^*\right) \notag \\
& +\frac{1}{2}\left(\tilde{\boldsymbol{\theta}}_{\tilde{\mathcal{A}}}-\boldsymbol{\theta}_{\tilde{\mathcal{A}}}^*\right)^{\top} \bSigma_{1,\tilde{\mathcal{A}}\times \tilde{\mathcal{A}}}\left(\tilde{\boldsymbol{\theta}}_{\tilde{\mathcal{A}}}-\boldsymbol{\theta}_{\tilde{\mathcal{A}}}^*\right)+\left(-\boldsymbol{\theta}_{\mathcal{I}_1}^*\right)^{\top} \bSigma_{1,\tilde{\mathcal{I}}_1 \times \tilde{\mathcal{A}}}\left(\tilde{\boldsymbol{\theta}}_{\tilde{\mathcal{A}}}-\boldsymbol{\theta}_{\tilde{\mathcal{A}}}^*\right) \notag \\
& +\frac{1}{2}\left(-\boldsymbol{\theta}_{\tilde{\mathcal{I}}_1}^*\right)^{\top} \bSigma_{1,\tilde{\mathcal{I}}_1 \times \tilde{\mathcal{I}}_1}\left(-\boldsymbol{\theta}_{\tilde{\mathcal{I}}_1}^*\right) . \label{eqtay}
\end{align}
Following up with the inequalities:
\begin{align}
l(\tilde{\boldsymbol{\theta}})-l\left(\boldsymbol{\theta}^*\right) &\leq \frac{M_s}{2}\left\|\boldsymbol{\theta}_{\tilde{\mathcal{I}}_1}^*\right\|_2^2 + \frac{\nu_s}{2}\left\|\boldsymbol{\theta}_{\tilde{\mathcal{I}}_1}^*\right\|_2\left\|\tilde{\boldsymbol{\theta}}_{\tilde{\mathcal{A}}}-\boldsymbol{\theta}_{\tilde{\mathcal{A}}}^*\right\|_2 + \frac{M_s}{2}\left\|\tilde{\boldsymbol{\theta}}_{\tilde{\mathcal{A}}}-\boldsymbol{\theta}_{\tilde{\mathcal{A}}}^*\right\|_2^2 \notag \\
& +\left|\left(\nabla_{\tilde{\mathcal{I}}_1} l(\boldsymbol{\theta^*})\right)^{\top}\left(-\boldsymbol{\theta}_{\tilde{\mathcal{I}}_1}^*\right)\right| + \left|\left(\nabla_{\tilde{\mathcal{A}}} l(\boldsymbol{\theta^*})\right)^{\top}\left(\tilde{\boldsymbol{\theta}}_{\tilde{\mathcal{A}}}-\boldsymbol{\theta}_{\tilde{\mathcal{A}}}^*\right)\right| \notag \\
& \leq \left(\frac{M_s}{2}+\frac{\nu_s^2}{2m_s}+\frac{M_s \nu_s^2}{2 m_s^2}\right)\left\|\boldsymbol{\theta}_{\mathcal{I}_{12} \cup \mathcal{A}_{12}}^*\right\|_2^2 + \varepsilon(\tilde{\mathcal{A}}).  \label{eq3.1.2}
\end{align}
where,
$$
\begin{aligned}
\varepsilon(\tilde{\mathcal{A}}) 
&= \left(\frac{M_s \nu_s}{m_s^2} + \frac{2 \nu_s}{m_s}\right) \left\|\boldsymbol{\theta}_{\mathcal{I}_{12} \cup \mathcal{A}_{12}}^*\right\|_2 \cdot \left\|\nabla_{\tilde{\mathcal{A}}} l\left(\boldsymbol{\theta}^*\right)\right\|_2 
 + \left\|\boldsymbol{\theta}_{\mathcal{I}_{12} \cup \mathcal{A}_{12}}^*\right\|_2 \cdot \left\|\nabla_{\mathcal{I}_{12} \cup \mathcal{A}_{12}} l\left(\boldsymbol{\theta}^*\right)\right\|_2 \\
& + \left(\frac{M_s}{2 m_s^2} + \frac{1}{m_s}\right) \left\|\nabla_{\mathcal{\tilde 
 A}} l\left(\boldsymbol{\theta}^*\right)\right\|_2 .
\end{aligned}
$$
Similarly, we can also derive:
\begin{align*}
l(\hat{\boldsymbol{\theta}}) & = l(\boldsymbol{\theta^*}) + \left(\left.\nabla_{\mathcal{\hat A}} l(\boldsymbol{\theta^*})\right)\right.^{\top} \left(\hat{\boldsymbol{\theta}}_{\hat{\mathcal{A}}} - \boldsymbol{\theta}_{\hat{\mathcal{A}}}^*\right) + \left(\left.\nabla_{\mathcal{I}_1} l(\boldsymbol{\theta^*})\right)\right.^{\top} \left(-\boldsymbol{\theta}_{\mathcal{I}_1}^*\right) \\
& + \frac{1}{2}\left(\hat{\boldsymbol{\theta}}_{\hat{\mathcal{A}}} - \boldsymbol{\theta}_{\hat{\mathcal{A}}}^*\right)^{\top} \bSigma_{1,\mathcal{\hat A}\times \mathcal{\hat A}} \left(\hat{\boldsymbol{\theta}}_{\hat{\mathcal{A}}} - \boldsymbol{\theta}_{\hat{\mathcal{A}}}^*\right) + \left(-\boldsymbol{\theta}_{\mathcal{I}_1}^*\right)^{\top} \bSigma_{1,\mathcal{I}_1\times \mathcal{\hat A}} \left(\hat{\boldsymbol{\theta}}_{\hat{\mathcal{A}}} - \boldsymbol{\theta}_{\hat{\mathcal{A}}}^*\right) \\
& + \frac{1}{2}\left(-\boldsymbol{\theta}_{\mathcal{I}_1}^*\right)^{\top} \bSigma_{1,\mathcal{I}_1 \times \mathcal{I}_1} \left(-\boldsymbol{\theta}_{\mathcal{I}_1}^*\right) .
\end{align*}
and 
\begin{align}
l(\hat{\boldsymbol{\theta}}) - l(\boldsymbol{\theta^*}) \geq & \frac{m_s}{2} \left\|\boldsymbol{\theta}_{\mathcal{I}_1}^*\right\|_2^2 - \frac{\nu_s}{m_s} \left\|\boldsymbol{\theta}_{\mathcal{I}_1}^*\right\|_2 \left\|\hat{\boldsymbol{\theta}}_{\hat{\mathcal{A}}} - \boldsymbol{\theta}_{\hat{\mathcal{A}}}^*\right\|_2 
 - \frac{M_s}{2m_s} \left\|\hat{\boldsymbol{\theta}}_{\hat{\mathcal{A}}} - \boldsymbol{\theta}_{\hat{\mathcal{A}}}^*\right\|_2^2  \notag \\
& - \left|\left(\nabla_{\mathcal{I}_1} l(\boldsymbol{\theta^*})\right)^{\top} \left(-\boldsymbol{\theta}_{\mathcal{I}_1}^*\right)\right| 
 - \left|\left(\nabla_{\mathcal{A}} l(\boldsymbol{\theta^*})\right)^{\top} \left(\hat{\boldsymbol{\theta}}_{\hat{\mathcal{A}}} - \boldsymbol{\theta}_{\hat{\mathcal{A}}}^*\right)\right| \notag \\
\geq & \left(\frac{m_s}{2} - \frac{\nu_s^2}{m_s} - \frac{M_s \nu_s^2}{2 m_s^2}\right) \left\|\boldsymbol{\theta}_{\mathcal{I}_1}^*\right\|_2^2 - \varepsilon(\hat{\mathcal{A}}). \label{eq3.1.3}
\end{align}
By Assumption (A4), we have
\begin{align}
& (1-\Delta)\left(\frac{m_s}{2}-\frac{\nu_s^2}{m_s}-\frac{M_s \nu_s^2}{2 m_s^2}\right)\left\|\boldsymbol{\theta}_{\mathcal{I}_1}^*\right\|_2^2 \notag \\
& \geq \frac{1}{2}\left(\frac{M_s}{2}+\frac{\nu_s^2}{m_s}+\frac{M_s \nu_s^2}{2 m_s^2}\right)\left[\left(2+\Delta+\frac{2 \nu_s}{m_s}\right)^2+(2+\Delta)^2\right]\left(\frac{\nu_s}{m_s}\right)^2\left\|\boldsymbol{\theta}_{\mathcal{I}_1}^*\right\|_2^2, \label{eq3.1.4}
\end{align}
and similarly, we can prove that
\begin{align}
& \mathbb{P}\left(\varepsilon(\tilde{\mathcal{A}})+\varepsilon(\hat{\mathcal{A}}) \geq \Delta\left(\frac{m_s}{2}-\frac{\nu_s^2}{m_s}-\frac{M_s \nu_s^2}{2 m_s^2}\right)\left\|\boldsymbol{\theta}_{\mathcal{I}_1}^*\right\|_2^2\right) \notag \\
\leq &  c_1 p \exp \left\{- C_2 \Delta^2\left(\frac{m_s}{2}-\frac{\nu_s^2}{m_s}-\frac{M_s \nu_s^2}{2 m_s^2}\right)^2\left\|\boldsymbol{\theta}_{\mathcal{I}_1}^*\right\|_2^2 /\left(s \sigma^2\right)\right\}  \notag\\
 \leq & \delta_3, \label{eq3.1.11}
\end{align}
where $C$ is some constant. \\
Therefore, with probability at least $1-\delta_1-\delta_2-\delta_3$, we have
$$
\begin{aligned}
l(\hat{\boldsymbol{\theta}})-l(\tilde{\boldsymbol{\theta}}) & >\left(1-\rho_s\right)(1-\Delta)\left(\frac{m_s }{2}-\frac{n \nu_s^2}{m_s }-\frac{ M_s \nu_s^2}{2 m_s^2 }\right)\left\|\boldsymbol{\theta}_{\mathcal{I}_1}^*\right\|_2^2 \\
& >C_2\left(1-\rho_s\right)(1-\Delta)  \min _{j \in \mathcal{A}^*}\left|\theta_j^*\right|^2 .
\end{aligned}
$$
Then by Assumption (A5) and (A6), for sufficiently large $n$, we have
\[
l(\hat{\boldsymbol{\theta}})-l(\tilde{\boldsymbol{\theta}})>\tau_s,
\]
which leads to a contradiction with $\mathcal{I}_1 \neq \varnothing$.

Throughout the proof, we have posited two critical assertions. Assuming the validity of these assertions, we've established the following result:
\[
\mathbb{P}\left(\hat{\mathcal{A}} \supseteq \mathcal{A}^*\right) \geq 1-\delta_1-\delta_2-\delta_3.
\]
This solidifies our main conclusion. It is crucial to recognize that the above result hinges critically on the two aforementioned assertions. In the ensuing discussion, we will delve deeper into these assertions and rigorously prove their validity.

1. 
\begin{equation*}
    \| \boldsymbol{\theta}^*_{\mathcal{A}_{12}}  \|_2 \le (2+\Delta) \frac{\nu_s}{m_s} \| \boldsymbol{\theta}^*_{\mathcal{I}_{1}} \|_2.
\end{equation*} 
Since $\left|\mathcal{A}_{12}\right|+\left|\mathcal{A}_{22}\right|=k_0=\left|\mathcal{I}_1\right| \leq\left|\mathcal{A}_2\right|=\left|\mathcal{A}_{21}+\mathcal{A}_{12}\right|$, we have $\left|\mathcal{A}_{12}\right| \leq\left|\mathcal{A}_{21}\right|$. By the definition of $\mathcal{S}_1$, we have
$$
\max _{j \in \mathcal{A}_{12}} \xi_j \leq \min _{j \in \mathcal{A}_{21}} \xi_j.
$$
Then
$$
\max _{j \in \mathcal{A}_{12}}\left(\hat{\theta}_j\right)^2 \leq \min _{j \in \mathcal{A}_{21}}\left(\hat{\theta}_j\right)^2 .
$$
Thus, we have
$$
\frac{1}{\sqrt{\left|\mathcal{A}_{12}\right|}}\left\|\hat{\boldsymbol{\theta}}_{\mathcal{A}_{12}}\right\|_2 \leq  \frac{1}{\sqrt{\left|\mathcal{A}_{21}\right|}}\left\|\hat{\boldsymbol{\theta}}_{\mathcal{A}_{21}}\right\|_2.
$$
As established in (\ref{eq3.1.0}),
$$
\|\hat \btheta_{\hat \cA}-\btheta^*_{\hat \cA}\|_2 \le 
\|  \bSigma_{1,\hat \cA \times \hat A}^{-1} (\nabla_{\hat \cA} l(\btheta^*)+ \bSigma_{1,\hat \cA \times  \cI_1} \boldsymbol{\theta}^*_{\hat \cI_1} )\|_2 
\le \frac{1}{m_s} \| \nabla_{\hat \cA} l(\btheta^*)\|_2+ \frac{\nu_s}{m_s}\| \boldsymbol{\theta}^*_{ \cI_1} \|_2.
$$
Therefore,
\begin{equation}
    \left\|\hat{\boldsymbol{\theta}}_{\mathcal{A}_{12}}\right\|_2 \geq \left\|\boldsymbol{\theta}_{\mathcal{A}_{12}}^*\right\|_2 - \frac{1}{m_s}\left\|\nabla_{\mathcal{\hat A}} l\left(\boldsymbol{\theta}^*\right)\right\|_2 - \frac{\nu_s}{m_s}\left\|\boldsymbol{\theta}_{\mathcal{I}_1}^*\right\|_2, \label{eq3.1.5}
\end{equation}
and
\begin{equation}
    \left\|\hat{\boldsymbol{\theta}}_{\mathcal{A}_{21}}\right\|_2 \leq \left\|\boldsymbol{\theta}_{\mathcal{A}_{21}}^*\right\|_2 + \frac{1}{m_s}\left\|\nabla_{\mathcal{\hat  A}} l\left(\boldsymbol{\theta}^*\right)\right\|_2 + \frac{\nu_s}{m_s}\left\|\boldsymbol{\theta}_{\mathcal{I}_1}^*\right\|_2.\label{eq3.1.6}
\end{equation}
Taking the difference between (\ref{eq3.1.5}) and (\ref{eq3.1.6}) yields:
\begin{align}
\left\|\boldsymbol{\theta}_{\mathcal{A}_{12}}^*\right\|_2 & \leq \left(1+ \sqrt{\frac{\left|\mathcal{A}_{12}\right|}{\left|\mathcal{A}_{21}\right|}}\right)\left(\frac{1}{m_s}\left\|\nabla_{\mathcal{\hat  A}} l\left(\boldsymbol{\theta}^*\right)\right\|_2 + \frac{\nu_s}{m_s}\left\|\boldsymbol{\theta}_{\mathcal{I}_1}^*\right\|_2\right) \notag \\
& \leq 2\left(\frac{1}{m_s}\left\|\nabla_{\mathcal{\hat  A}} l\left(\boldsymbol{\theta}^*\right)\right\|_2 + \frac{\nu_s}{m_s}\left\|\boldsymbol{\theta}_{\mathcal{I}_1}^*\right\|_2\right). \label{eq3.1.7}
\end{align}
Recall that:
$$
\nabla l\left(\boldsymbol{\theta}^*\right) = -\frac{1}{N} \mathbf{X}^{\top} \boldsymbol{\epsilon} + \left(\boldsymbol{\Sigma} - \boldsymbol{\Sigma}_1\right) \boldsymbol{\delta}.
$$
Then, the $\ell_2$ norm of this gradient can be bounded as follows:
$$
\|\nabla l\left(\boldsymbol{\theta}^*\right)\|_2 \leq \frac{1}{N}\|\mathbf{X}^{\top}\boldsymbol{\epsilon}\|_2 + \|\left(\boldsymbol{\Sigma} - \boldsymbol{\Sigma}_1\right) \boldsymbol{\delta}\|_2.
$$
For any $t>0$ and an index set $\mathcal{B}$  with $|\mathcal{B}| \leq s$, we have:
$$
\begin{aligned}
\mathbb{P}\left(\left\|\frac{1}{N} \mathbf{X}_{\mathcal{B}}^{\top} \boldsymbol{\epsilon}\right\|_2 \geq t\right) &\leq \mathbb{P}\left(\max _{|\mathcal{A}| \leq s}\left\|\frac{1}{\sqrt{N}} \mathbf{X}_{\mathcal{A}}^{\top} \boldsymbol{\epsilon}\right\|_2^2 \geq Nt^2\right) \\
&\leq \sum_{j=1}^p \mathbb{P}\left(\left\|\frac{1}{\sqrt{N}} \mathbf{X}_j^{\top} \boldsymbol{\epsilon}\right\|_2 \geq t \sqrt{\frac{N}{s}}\right) \\
&\leq 2p \exp \left\{-\frac{Nt^2}{s M_1^2 \sigma^2}\right\}.
\end{aligned}
$$
Then, we deduce that:
$$
\mathbb{P}\left(\frac{1}{N}\left\|\mathbf{X}_{\hat{\mathcal{A}}}^{\top} \boldsymbol{\epsilon}\right\|_2 \geq \frac{\Delta \nu_s}{4} \left\|\boldsymbol{\theta}_{\mathcal{I}_1}^*\right\|_2\right) \leq 2p \exp \left\{-\frac{N \Delta^2 \nu_s^2\left\|\boldsymbol{\theta}_{\mathcal{I}_1}^*\right\|_2^2}{4 s M_1^2 \sigma^2}\right\} \leq \frac{\delta_1}{2}.
$$
Given that: $\displaystyle
\|(\boldsymbol{\Sigma} - \boldsymbol{\Sigma}_1)\boldsymbol{\delta}\|_2 \lesssim \sqrt{\frac{s \log p}{N}}$
and 
$\displaystyle
\frac{1}{N}\left\|\mathbf{X}_{\hat{\mathcal{A}}}^{\top} \boldsymbol{\epsilon}\right\|_2 = O\left(\sqrt{\frac{s \log p}{N}}\right),
$
it follows that there exists a constant $k_1$ such that:
$$
\|(\boldsymbol{\Sigma} - \boldsymbol{\Sigma}_1)\boldsymbol{\delta}\|_2 \leq \frac{k_1}{N}\left\|\mathbf{X}_{\hat{\mathcal{A}}}^{\top} \boldsymbol{\epsilon}\right\|_2.
$$
Therefore,
$$
\mathbb{P}\left(\|(\boldsymbol{\Sigma} - \boldsymbol{\Sigma}_1)\boldsymbol{\delta}\|_2 \geq \frac{\Delta \nu_s}{4} \left\|\boldsymbol{\theta}_{\mathcal{I}_1}^*\right\|_2\right) \leq 2p \exp \left\{-\frac{k_1^2 N \Delta^2 \nu_s^2\left\|\boldsymbol{\theta}_{\mathcal{I}_1}^*\right\|_2^2}{4 s M_1^2 \sigma^2}\right\} \leq \frac{\delta_1}{2}.
$$
Thus, we can get
$$
\mathbb{P}\left(\left\|\boldsymbol{\theta}_{\mathcal{A}_{12}}^*\right\|_2 \leq (2+\Delta) \frac{\nu_s}{m_s }\left\|\boldsymbol{\theta}_{\mathcal{I}_1}^*\right\|_2\right) \geq 1-\delta_1.
$$

2. 
\begin{equation*}
    \| \boldsymbol{\theta}^*_{\mathcal{I}_{12}}  \|_2 \le 2(1+\Delta) (\nu_s+\frac{\nu_s}{m_s}) \| \boldsymbol{\theta}^*_{\mathcal{I}_{1}}  \|_2.
\end{equation*}
Since $\left|\mathcal{I}_{11}\right|+\left|\mathcal{I}_{21}\right|=\left|\mathcal{S}_2\right|=\left|\mathcal{I}_1\right|=\left|\mathcal{I}_{11}\right|+\left|\mathcal{I}_{12}\right|$, we have $\left|\mathcal{I}_{12}\right|=\left|\mathcal{I}_{21}\right|$. By the definition of $\mathcal{S}_2$, we have
$$
\min _{j \in \mathcal{I}_{21}}\left|\zeta_j\right| \geq \max _{j \in \mathcal{I}_{12}}\left|\zeta_j\right| .
$$
Thus,
$$
\min _{j \in \mathcal{I}_{21}}\hat{d}_j^2 \geq \max_{j \in \mathcal{I}_{12}} \hat{d}_j^2.
$$
Note that,
$$
\hat{\boldsymbol{d}}_{\hat{\mathcal{I}}} = \nabla_{\hat{\mathcal{I}}} l\left(\boldsymbol{\theta}^*\right) + \bSigma_{1,\hat{\mathcal{I}} \times \mathcal{I}_1} \left(-\boldsymbol{\theta}_{\mathcal{I}_1}^*\right) + \bSigma_{1,\hat{\mathcal{I}} \times \hat{\mathcal{A}}} \left(\hat{\boldsymbol{\theta}}_{\hat{\mathcal{A}}} - \boldsymbol{\theta}_{\hat{\mathcal{A}}}^*\right) .
$$
Therefore, we have
k
\begin{align}
  \left\|\hat{\boldsymbol{d}}_{\mathcal{I}_{12}}\right\|_2 & =\left\| \nabla_{\mathcal{I}_{12}} l\left(\boldsymbol{\theta}^*\right) + \bSigma_{1,\mathcal{I}_{12} \times \mathcal{I}_1} \left(-\boldsymbol{\theta}_{\mathcal{I}_1}^*\right) + \bSigma_{1,\mathcal{I}_{12} \times \hat{\mathcal{A}}} \left(\hat{\boldsymbol{\theta}}_{\hat{\mathcal{A}}} - \boldsymbol{\theta}_{\hat{\mathcal{A}}}^*\right)\right\|_2 \notag \\
  & \geq m_s\left\|\boldsymbol{\theta}_{\mathcal{I}_{12}}^*\right\|_2 - \left\| \nabla_{\mathcal{I}_{12}} l\left(\boldsymbol{\theta}^*\right) \right\|_2 - \frac{\nu_s}{m_s}\left\|\hat{\boldsymbol{\theta}}_{\hat{\mathcal{A}}} - \boldsymbol{\theta}_{\hat{\mathcal{A}}}^*\right\|_2 \notag \\
  & \geq m_s\left\|\boldsymbol{\theta}_{\mathcal{I}_{12}}^*\right\|_2 - \left\| \nabla_{\mathcal{I}_{12}} l\left(\boldsymbol{\theta}^*\right) \right\|_2 - \frac{\nu_s}{m_s^2} \left\| \nabla_{\hat{\mathcal{A}}} l\left(\boldsymbol{\theta}^*\right) \right\|_2 - \frac{\nu_s^2}{m_s^2} \left\|\boldsymbol{\theta}_{\mathcal{I}_1}^*\right\|_2, \label{eq3.1.8}
\end{align}
and

\begin{align}
  \left\|\hat{\boldsymbol{d}}_{\mathcal{I}_{21}}\right\|_2 & = \left\| \nabla_{\mathcal{I}_{21}} l\left(\boldsymbol{\theta}^*\right) + \bSigma_{1,\mathcal{I}_{21} \times \mathcal{I}_1} \left(-\boldsymbol{\theta}_{\mathcal{I}_1}^*\right) + \bSigma_{1,\mathcal{I}_{21} \times \hat{\mathcal{A}}} \left(\hat{\boldsymbol{\theta}}_{\hat{\mathcal{A}}} - \boldsymbol{\theta}_{\hat{\mathcal{A}}}^*\right) \right\|_2 \notag \\
  & \leq \left\| \nabla_{\mathcal{I}_{21}} l\left(\boldsymbol{\theta}^*\right) \right\|_2 + \frac{\nu_s}{m_s}\left\|\boldsymbol{\theta}_{\mathcal{I}_1}^*\right\|_2 + \frac{\nu_s}{m_s^2}\left\| \nabla_{\hat{\mathcal{A}}} l\left(\boldsymbol{\theta}^*\right) \right\|_2 \notag \\
  & \leq \left\| \nabla_{\mathcal{I}_{21}} l\left(\boldsymbol{\theta}^*\right) \right\|_2 + \frac{\nu_s}{m_s}\left\|\boldsymbol{\theta}_{\mathcal{I}_1}^*\right\|_2 + \frac{\nu_s}{m_s^2}\left\| \nabla_{\hat{\mathcal{A}}} l\left(\boldsymbol{\theta}^*\right) \right\|_2 + \frac{\nu_s^2}{m_s^2}\left\|\boldsymbol{\theta}_{\mathcal{I}_1}^*\right\|_2. \label{eq3.1.9}
\end{align}
Since $\left|\mathcal{I}_{12}\right| = \left|\mathcal{I}_{21}\right|$, $\left\|\hat{\boldsymbol{d}}_{\mathcal{I}_{21}}\right\|_2 \geq\left\|\hat{\boldsymbol{d}}_{\mathcal{I}_{12}}\right\|_2$, we have
\begin{align}
& 2\nu_s \left\|\boldsymbol{\theta}_{\mathcal{I}_1}^*\right\|_2 + \frac{2 \nu_s^2}{m_s}\left\|\boldsymbol{\theta}_{\mathcal{I}_1}^*\right\|_2 \notag \\
 \geq & m_s\left\|\boldsymbol{\theta}_{\mathcal{I}_{12}}^*\right\|_2 - \frac{2 \nu_s}{m_s}\left\| \nabla_{\hat{\mathcal{A}}} l\left(\boldsymbol{\theta}^*\right) \right\|_2 - \left\| \nabla_{\mathcal{I}_{12}} l\left(\boldsymbol{\theta}^*\right) \right\|_2 - \left\| \nabla_{\mathcal{I}_{21}} l\left(\boldsymbol{\theta}^*\right) \right\|_2. \label{eq3.1.10}
\end{align}
By assumption (A2), We have
$$
\begin{aligned}
& \mathbb{P}\left(m_s\left\| \nabla_{\mathcal{I}_{12}} l\left(\boldsymbol{\theta}^*\right) \right\|_2 \geq \frac{\Delta}{3} \frac{\nu_s}{m_s}\left\|\boldsymbol{\theta}_{\mathcal{I}_1}^*\right\|_2\right) \nonumber 
 \quad \leq c_1 p \exp \left\{-\frac{  c_2 \Delta^2 \nu_s^2\left\|\boldsymbol{\theta}_{\mathcal{I}_1}^*\right\|_2^2}{9 m_s^2 s \sigma^2}\right\} \leq \delta_2 / 3, \nonumber \\
& \mathbb{P}\left(\frac{2 \nu_s}{m_s^3}\left\| \nabla_{\hat{\mathcal{A}}} l\left(\boldsymbol{\theta}^*\right) \right\|_2 \geq \frac{\Delta}{3} \frac{\nu_s}{m_s}\left\|\boldsymbol{\theta}_{\mathcal{I}_1}^*\right\|_2\right) \nonumber 
 \quad \leq c_1 p \exp \left\{-\frac{  \Delta^2 c_2 m_s^2\left\|\boldsymbol{\theta}_{\mathcal{I}_1}^*\right\|_2^2}{36 s \sigma^2}\right\} \leq \delta_2 / 3, \nonumber \\
& \mathbb{P}\left(\left\| \nabla_{\mathcal{I}_{21}} l\left(\boldsymbol{\theta}^*\right) \right\|_2 \geq \frac{\Delta}{3} \frac{\nu_s}{m_s}\left\|\boldsymbol{\theta}_{\mathcal{I}_1}^*\right\|_2\right) \nonumber 
\quad \leq c_1 p \exp \left\{-\frac{  c_2 \Delta^2 \nu_s^2\left\|\boldsymbol{\theta}_{\mathcal{I}_1}^*\right\|_2^2}{9 m_s s \sigma^2}\right\} \leq \delta_2 / 3 .
\end{aligned}
$$
Consequently, we have
$$
\mathbb{P}\left(\nu_s \left(2+\frac{2 \nu_s}{m_s^2}+\Delta\right)  \left\|\boldsymbol{\theta}_{\mathcal{I}_1}^*\right\|_2 \geq m_s\left\|\boldsymbol{\theta}_{\mathcal{I}_{12}}^*\right\|_2\right) \geq 1-\delta_2 .
$$
Therefore, we have successfully established the two initial assertions that underpin our main result.
\end{proof}

\subsection{Proof of Theorem 3.2}
In the following discussion, let us denote the output of Algorithm 1 for a given support size $s$ as $(\hat{\boldsymbol{\theta}}^s, \hat{\mathcal{A}}^s)$. Specifically, let $s^*$ represent the true sparsity. Consider the function $f(\btheta)$ defined as:
$$
f(\btheta) = \frac{1}{2N} \|\bY - \bX\btheta\|_2^2 = \frac{1}{m} \sum_{k=1}^m f_k(\btheta) = \frac{1}{m} \sum_{k=1}^m \frac{1}{2n} \|\bY_k - \bX_k\btheta\|_2^2.
$$
Using the inequality $1 - \frac{1}{x} \leq \log(x) \leq x - 1$ for $x > 0$, for any $\btheta_1$ and $\btheta_2$, we have:
\begin{equation}
\frac{f(\btheta_1) - f(\btheta_2)}{f(\btheta_1)} \leq \log \frac{f(\btheta_1)}{f(\btheta_2)} \leq \frac{f(\btheta_1) - f(\btheta_2)}{f(\btheta_2)}. \label{eqn-3.2.1}
\end{equation}
By Theorem 3.1, with probability $1 - O(p^{-\alpha})$, we obtain:
$$
\hat{\mathcal{A}}^s \supseteq \mathcal{A}^*, \text{ for } s \geq s^*.
$$
Let $\hat{\mathcal{A}}^s = \mathcal{A}^* \cup \mathcal{B}^s$. According to the supplementary material from \cite{Zhu2020Polynomial}, specifically equation (15) in the proof of Theorem 4, we have:
$$
f_k(\hat{\boldsymbol{\theta}}^{s^*}) - f_k(\hat{\boldsymbol{\theta}}^s) \leq \frac{\sigma^2 |\mathcal{B}^s|}{2n}(1 + \alpha) \log(2p), \text{ for } k = 1, \ldots, m.
$$
Summing over $k$, we get:
\begin{equation}
    f(\hat{\boldsymbol{\theta}}^{s^*}) - f(\hat{\boldsymbol{\theta}}^s) \leq \frac{\sigma^2 |\mathcal{B}^s|}{2N}(1 + \alpha) \log(2p). \label{eqn-3.2.2}
\end{equation}
Similarly,  with probability $1 - O(p^{-\alpha})$, we have:
\begin{align*}
f_k(\hat{\boldsymbol{\theta}}^s) & = \frac{1}{2n} \|\bY_k - \left({\bX_k}\right)_{\hat{\mathcal{A}}^s} \hat{\boldsymbol{\theta^s}}_{\hat{\mathcal{A}}^s}\|_2^2 \\
& \geq \frac{1}{2n} \|\boldsymbol{\epsilon}_k\|_2^2 - \frac{1}{2n} \left({\bX_k}\right)_{\hat{\mathcal{A}}^s} (\hat{\boldsymbol{\theta}^s}_{\hat{\mathcal{A}}^s} - \boldsymbol{\theta}_{\hat{\mathcal{A}}^s}^*)\|_2^2 \\
& \geq \frac{\sigma^2}{2} \left(1 - \frac{1}{2n} \alpha \log(2p)\right) - \frac{\sigma^2 s}{2n}(1 + \alpha) \log(2p) \\
& > 0.
\end{align*}
Summing over $k$:
\begin{equation}
    f(\hat{\boldsymbol{\theta}}^s) \geq \frac{\sigma^2}{2} \left(1 - \frac{1}{2N} \alpha \log(2p)\right) - \frac{\sigma^2 s}{2N}(1 + \alpha) \log(2p).\label{eqn-3.2.3}
\end{equation}
Combining equations (\ref{eqn-3.2.1}), (\ref{eqn-3.2.2}), and (\ref{eqn-3.2.1}), with probability $1 - O(p^{-\alpha})$, we obtain:
$$
\log \frac{f(\hat{\boldsymbol{\theta}}^{s^*})}{f(\hat{\boldsymbol{\theta}}^s)} \leq \frac{\frac{\sigma^2 |\mathcal{B}^s|}{N}(1 + \alpha) \log(2p)}{\frac{\sigma^2}{2} \left(1 - \frac{1}{2N} \alpha \log(2p)\right) - \frac{\sigma^2 s}{2N}(1 + \alpha) \log(2p)}.
$$
Consequently:
$$
\begin{aligned}
\operatorname{GIC}(\hat{\btheta}^{s^*}) - \operatorname{GIC}(\hat{\btheta}^s) & = N \log \frac{f(\hat{\btheta}^{s^*})}{f(\hat{\btheta}^s)} - |\mathcal{B}^s| \log(p) \log \log N \\
& \leq O\left(|\mathcal{B}^s| \log(2p)\right) - |\mathcal{B}^s| \log(p) \log \log N \\
& < 0
\end{aligned}
$$
for sufficiently large $N$.

On the other hand, consider the case when $ s < s^* $. Denote
$$
\begin{aligned}
& \mathcal{A}_1^s=\hat{\mathcal{A}}^s \cap \mathcal{A}^*, \mathcal{A}_2^s=\hat{\mathcal{A}}^s \cap \mathcal{I}^*, \\
& \mathcal{I}_1^s=\left(\hat{\mathcal{A}}^s\right)^c \cap \mathcal{A}^*, \mathcal{I}_2^s=\left(\hat{\mathcal{A}}^s\right)^c \cap \mathcal{I}^* .
\end{aligned}
$$
Similar to (\ref{eqtay}), expand $ f(\boldsymbol{\theta})$ around $\boldsymbol{\theta}^* $:
$$
\begin{aligned}
f\left(\hat{\boldsymbol{\theta}}^s\right)=& f\left(\boldsymbol{\theta}^*\right)+ \left\{\nabla_{\hat{\mathcal{A}}^s} f(\boldsymbol{\theta}^*)\right\}^{\top}\left(\hat{\boldsymbol{\theta}}_{\hat{\mathcal{A}}^s}-\boldsymbol{\theta}_{\hat{\mathcal{A}}^s}^*\right)+\left\{\nabla_{\mathcal{I}_1^s} f(\boldsymbol{\theta}^*)\right\}^{\top}\left(-\boldsymbol{\theta}_{\mathcal{I}_1^s}^*\right) \\
& + \frac{1}{2}\left(\hat{\boldsymbol{\theta}}_{\hat{\mathcal{A}}^s}-\boldsymbol{\theta}_{\hat{\mathcal{A}}^s}^*\right)^{\top} \bSigma_{\hat{\mathcal{A}}^s\times \hat{\mathcal{A}}^s}\left(\hat{\boldsymbol{\theta}}_{\hat{\mathcal{A}}^s}-\boldsymbol{\theta}_{\hat{\mathcal{A}}^s}^*\right) 
 + \left(-\boldsymbol{\theta}_{\mathcal{I}_1^s}^*\right)^{\top} \bSigma_{\mathcal{I}_1^s\times \hat{\mathcal{A}}^s}\left(\hat{\boldsymbol{\theta}}_{\hat{\mathcal{A}}^s}-\boldsymbol{\theta}_{\hat{\mathcal{A}}^s}^*\right)\\
& +\frac{1}{2}\left(-\boldsymbol{\theta}_{\mathcal{I}_1^s}^*\right)^{\top} \bSigma_{\mathcal{I}_1^s\times \mathcal{I}_1^s}\left(-\boldsymbol{\theta}_{\mathcal{I}_1^s}^*\right) .
\end{aligned}
$$
Similar to (\ref{eq3.1.4}) and (\ref{eq3.1.11}), we can obtain that with probability $1-O\left(p^{-\alpha}\right)$, for some $0<\Delta<\frac{1}{2}$:
\begin{equation}
    f\left(\hat{\boldsymbol{\theta}}^s\right)-f\left(\boldsymbol{\theta}^*\right) \geq (1-\Delta)\left(\frac{m_s}{2}-\frac{\nu_s^2}{m_s}-\frac{\nu_s^2 M_s}{2 m_s^2}\right)\left\|\boldsymbol{\theta}_{\mathcal{I}_1^s}^*\right\|_2^2.\label{eq3.3.1}
\end{equation}
and 
$$
\left\|\hat{\boldsymbol{\theta}}_{\mathcal{A}^*}^*-\boldsymbol{\theta}_{\mathcal{A}^*}^*\right\|_2 \leq  \frac{1}{m_s}\left\|\nabla_{\mathcal{A}^*}f(\boldsymbol{\theta^*})\right\|_2 .
$$
Then
\begin{align}
f\left(\hat{\boldsymbol{\theta}}^{s^*}\right)-f\left(\boldsymbol{\theta}^*\right)= & \nabla_{\mathcal{A}^*} f(\boldsymbol{\theta^*})^{\top}\left(\hat{\boldsymbol{\theta}}_{\mathcal{A}^*}^*-\boldsymbol{\theta}_{\mathcal{A}^*}^*\right) \notag \\
& +\frac{1}{2}\left(\hat{\boldsymbol{\theta}}_{\mathcal{A}^*}^*-\boldsymbol{\theta}_{\mathcal{A}^*}^*\right)^{\top} \bSigma_{ \mathcal{A}^*\times\mathcal{A}^*}\left(\hat{\boldsymbol{\theta}}_{\mathcal{A}^*}^*-\boldsymbol{\theta}_{\mathcal{A}^*}^*\right) \notag\\
\leq & \left(\frac{1}{m_s}  +\frac{M_s}{m_s^2}\right)\left\|\nabla_{\mathcal{A}^*}f(\boldsymbol{\theta^*})\right\|_2^2 . \label{eq3.3.2}
\end{align}
It can be shown that
$$
\mathbb{P}\left((\frac{1}{m_s}  +\frac{M_s}{m_s^2}) \left\|\nabla_{\mathcal{A}^*} f (\boldsymbol{\theta^*})\right\|_2^2 \geq \Delta\left(\frac{m_s}{2}-\frac{\nu_s^2}{m_s}-\frac{\nu_s^2 M_s}{m_s^2}\right)\left\|\boldsymbol{\theta}_{\mathcal{I}_1^s}^*\right\|_2^2\right) \leq \delta .
$$
Then with probability at least $1-\delta$,
\begin{equation}
    f\left(\hat{\boldsymbol{\theta}}^s\right)-f\left(\hat{\btheta}^{s^*}\right) \geq (1-2 \Delta)\left(\frac{m_s}{2}-\frac{\nu_s^2}{m_s}-\frac{\nu_s^2 M_s}{m_s^2}\right)\left\|\boldsymbol{\theta}_{\mathcal{I}_1^s}^*\right\|_2^2.
\end{equation}
Consequently,
$$
\begin{aligned}
\text{GIC}\left(\hat{\boldsymbol{\theta}}^s\right)- \text{GIC}\left(\hat{\btheta}^{s^*}\right) & =N \log \frac{f(\hat\btheta^s)}{f(\hat{\btheta}^{s^*})} -\left(s^*-s\right) \log p \log \log N \\
& \geq  N \frac{f(\hat\btheta^s)-f(\hat{\btheta}^{s^*})}{f(\hat\btheta^s)} -\left(s^*-s\right) \log p \log \log N \\
& \geq NO\left(\left\|\boldsymbol{\theta}_{\mathcal{I}_1^s}^*\right\|_2^2\right)-s^* \log p \log \log N\\
& >0.
\end{aligned}
$$

Therefore, information criterion $\text{GIC}(\hat{\boldsymbol{\theta}}^s)$ attains minimum at $\hat{\boldsymbol{\theta}}^{s^*}$ with probability $1-p^{-\alpha}$.

\subsection{Proof of Theorem 3.3}
Given that $\btheta^k$ denotes the least squares estimate restricted to the active set for the $k$th machine, let $\hat{\btheta} = \frac{1}{m} \sum_{k=1}^m \btheta^k$. According to theorem 4 from \cite{Zhu2020Polynomial}, $\btheta^k$ is an unbiased estimate, i.e., $\mathbb{E}[\btheta^k] = \btheta^*$.
By simple algebraic manipulation, we can derive that:
\begin{align}
\mathbb{E}\left[\left\|\hat{\btheta}-\btheta^*\right\|_2^2\right] & =\mathbb{E}\left[\left\|\frac{1}{m} \sum_{k=1}^m \btheta^k-\btheta^*\right\|_2^2\right]  \notag \notag\\
& =\frac{1}{m^2} \sum_{k=1}^m \mathbb{E}\left[\left\|\btheta^k-\btheta^*\right\|_2^2\right]+\frac{1}{m^2} \sum_{k \neq j} \mathbb{E}\left[\left\langle\btheta^k-\btheta^*, \btheta^j-\btheta^*\right\rangle\right] \notag\\
& \leq \frac{1}{m} \mathbb{E}\left[\left\|\btheta^1-\btheta^*\right\|_2^2\right]+\frac{m(m-1)}{m^2}\left\|\mathbb{E}\left[\btheta^1-\btheta^*\right]\right\|_2^2 \notag\\
& \leq \frac{1}{m} \mathbb{E}\left[\left\|\btheta^1-\btheta^*\right\|_2^2\right]+\left\|\mathbb{E}\left[\btheta^1-\btheta^*\right]\right\|_2^2  \notag\\
& \leq \frac{1}{m} \mathbb{E}\left[\left\|\btheta^1-\btheta^*\right\|_2^2\right]. 
\label{eqt3.3.1}
\end{align}
Upon recovering  the active set, the OLS estimator is subject to the following bound \citep{Raskutti2011minimax, Zhu2020Polynomial}:
$$
\max _{\btheta^* \in \mathcal{B}\left(s^*\right)}\left\|\btheta^1-\btheta^*\right\|_2^2  \le C_5 \frac{s^* \log \left(p / s^*\right)}{n}
$$
with probability at least $1-O\left(p^{-\alpha}\right)-e^{-s^* \log \left(p /\left(e \cdot s^*\right)\right)}$ for some $\alpha>0$, where $\mathcal{B}\left(s^*\right)$ represents the $\ell_0$-ball defined as $\{\btheta \in \mathbb{R}^p:\|\btheta\|_0 \leq s^*\}$, and $C_5$ is a constant.
Therefore, by taking the expectation, there exists a constant $C_6$ such that:
\begin{equation}
    \mathbb{E} \left[\left\|\btheta^1 -\btheta^*\right\|_2^2 \right] \le C_6 \frac{s^* \log \left(p / s^*\right)}{n}. \label{eqt3.3.2}
\end{equation}
Substituting the  equation (\ref{eqt3.3.2} into (\ref{eqt3.3.1}), we obtain:
\begin{equation}
    \mathbb{E}\left[\left\|\hat{\btheta}-\btheta^*\right\|_2^2\right] 
 \leq \frac{1}{m} \mathbb{E}\left[\left\|\btheta^1-\btheta^*\right\|_2^2\right]
\le C_6 \frac{s^* \log \left(p / s^*\right)}{mn} = C_6 \frac{s^* \log \left(p / s^*\right)}{N}.
\end{equation}

\bibliographystyle{agsm}

\bibliography{reference}